\documentclass{article}
\usepackage{PRIMEarxiv}
\usepackage{fancyhdr}       
\usepackage{comment}
\usepackage[square,numbers]{natbib}

\usepackage{makecell}
\usepackage{iitem}
\usepackage{hyperref}
\usepackage[figuresright]{rotating}
\usepackage{longtable}
\usepackage{multirow}
\usepackage{adjustbox}
\usepackage{enumitem}
\usepackage{tablefootnote}
\usepackage{pdflscape}   

\title{From Data to Action: Charting A Data-Driven Path to Combat Antimicrobial Resistance}

\author{Qian Fu\thanks{Corresponding author: Qian Fu (qian.fu@data61.csiro.au)}, ~Yuzhe Zhang, Yanfeng Shu, Ming Ding, Lina Yao, Chen Wang\\
CSIRO Data61\\
Australia\\
}

\begin{document}
\maketitle

\begin{abstract}
Antimicrobial-resistant (AMR) microbes are becoming increasingly common in healthcare as they make modern medicines ineffective and result in serious problems. 
The cause of AMR is considered related to antibiotic production either in a natural environment or by synthetic processes and the impact of antibiotics on bacteria evolution. However, it is challenging to quantify factors affecting AMR transmissions for effective decision-making.
As the data related to AMR has been increasingly collected, 
data-driven methods have been increasingly used and are promising in providing meaningful clues for identifying AMR causes and effective treatment methods. 
In this paper, 
we review AMR works from the data analytics and machine learning perspective, 
in an attempt to summarise the state-of-the-art and give insight into the problem space.
In more detail, 
we explore the diverse aspects of AMR, 
including surveillance, prediction, drug discovery, stewardship, and driver analysis. 
Then, we elaborate on the interaction of these aspects and the common data-related methodologies employed. 
For data handling, 
we have discussed sources, methods, and challenges in collecting and analyzing AMR-related data, 
while underlining the importance of standardization and interoperability. 
Further, 
this article surveys data analysis techniques, 
from statistical analysis to machine learning/deep learning,
illustrating their application in tackling AMR challenges. 
Mainly aiming at data challenges, including noises and biases introduced in data preparation (e.g., data cleaning and privacy-preserving) and modelling phases, 
the paper also highlights strategies (i.e., denoising and debiasing) for mitigating data challenges to improve AMR research performance and results in terms of fairness and robustness. 
In conclusion, 
the paper focuses on problems in the intersection of AMR and data science, 
stressing the need for interdisciplinary collaboration, 
especially arousing the awareness of noise and bias in data-driven approaches in the ongoing battle against AMR. 
It points towards promising paths for further exploration, innovation, robustness and fairness in AMR research.
\end{abstract}


\maketitle

\section{Introduction}\label{sec:intro}

\begin{figure}[htbp]
 \centering
  \includegraphics[width=\textwidth]{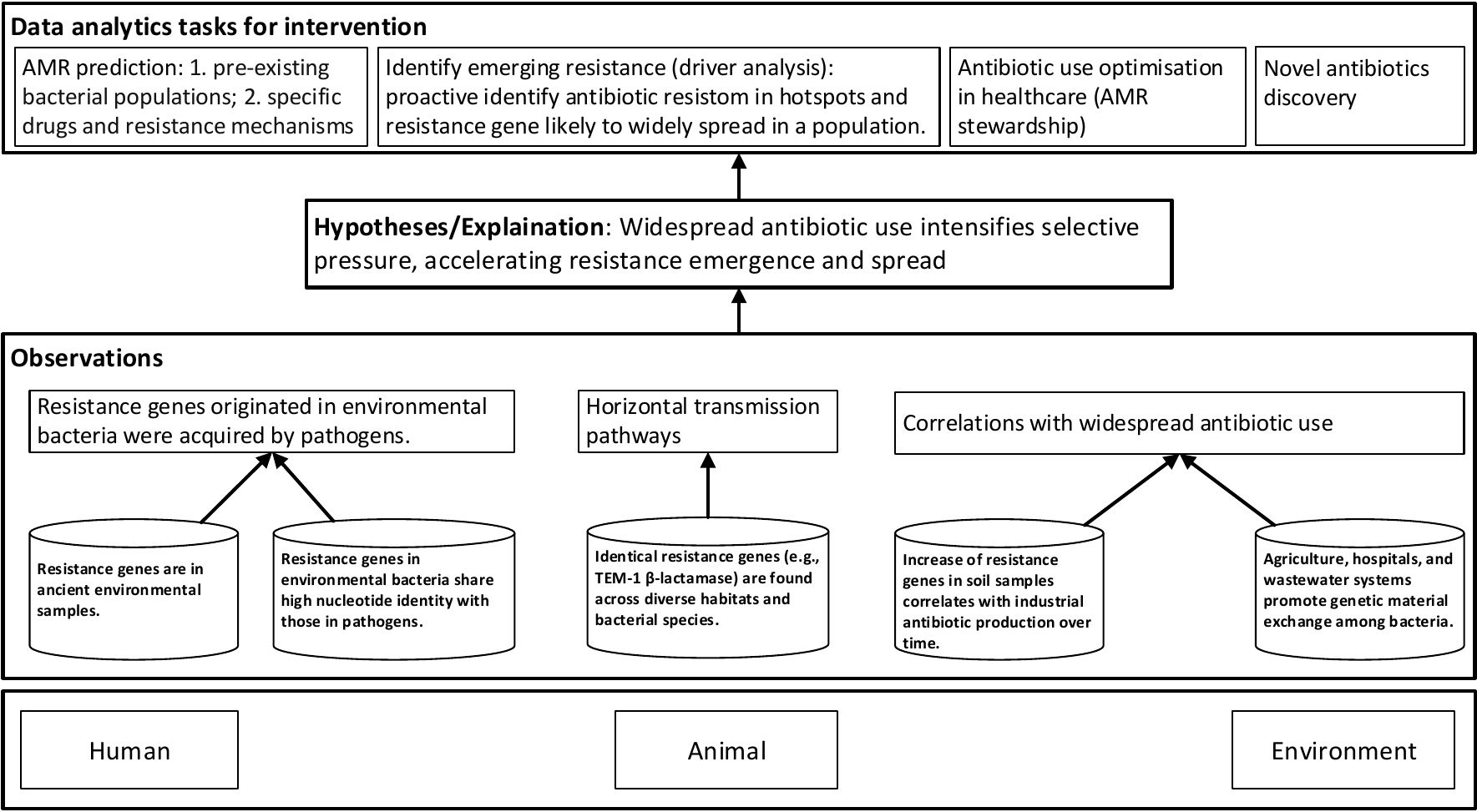}
  \caption{AMR transmissions among human, animal and environment: current observations and intervention gaps. Cylinders represent data supporting the AMR transmission pathways. Data analytics tasks are introduced for AMR intervention strategy development.}
  \label{fig:amr_rel}
\end{figure}

Antibiotics are often grouped by their mechanisms of action, 
such as blocking protein synthesis, disrupting folate biosynthesis, changing cell wall construction, compromising the cell membrane integrity and affecting DNA replication \cite{walsh2020antibiotics,crofts2017next}.
These antibiotics, whether created in labs or found in nature, serve as the primary defence against bacterial infections. However, bacteria employ a series of strategies in response to resist these antibiotics, including inactivating antibiotics through enzymatic degradation, altering the antibiotic target, modifying cell membrane permeability, and using efflux pumps to maintain intracellular antibiotic concentrations of antibiotics below inhibitory levels \cite{crofts2017next}.

Moreover, the gene transfer of antibiotic-resistant bacteria (ARB) further aggravates this challenge \cite{vikesland2019differential}. Resistance can be gained through vertical and horizontal gene transfers, of which the former is the transfer from parent to offspring, and the latter is the transfer of genetic material, including antimicrobial-resistant genes (ARGs), from cell to cell. These transfers happen within microbial communities and can even extend to diverse environments (e.g., human, animal, water and soil), spreading resistance.

Even though resistance genes exist in ancient environment samples, antimicrobial resistance (AMR) has become an urgent threat challenging global public health in recent years due to the importance of antibiotics in healthcare. The efficacy of antimicrobials, essential in combating bacterial infections, is harmed by the continuous evolution of ARB. Current research attributes the spread of AMR to the complex interaction among human, animal and environment~\cite{crofts2017next, booton2021one}, as shown in Fig.~\ref{fig:amr_rel}, evidenced by data collected from a variety of sources.
AMR surveillance serves as a major effort of data collection. It tracks the AMR and antimicrobial usage status in the bodies of humans, animals and the environment, and monitors the efficacy of strategies on all levels (local, national, and global). 

\begin{figure*}[htbp]
 \centering
  \includegraphics[width=1.0\textwidth]{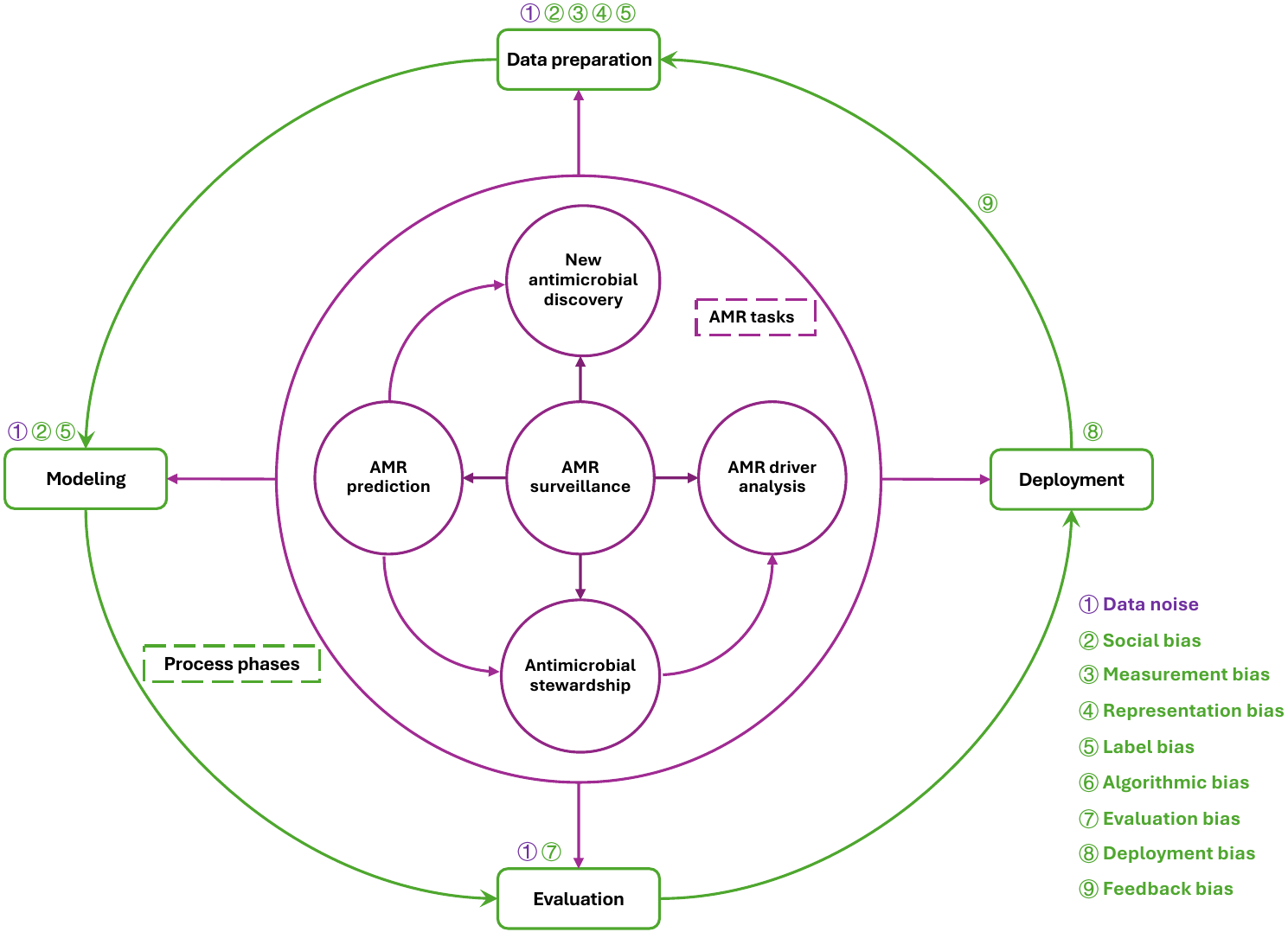}
  \caption{Process phases and potential data challenges of learning-based AMR tasks, where \textcircled{1} stands for data noise and \textcircled{2}-\textcircled{9} are 8 different types of data biases.}
  \label{fig:introduction_overview}
\end{figure*}

In the AMR field, various interconnected tasks work together to combat this pressing global health issue. In this paper, we investigate the following major tasks of AMR:
\begin{itemize}
\item AMR driver analysis: identifying and quantifying the factors that contribute to the emergence and dissemination of antimicrobial-resistant pathogens.
\item Antimicrobial stewardship: coordinated interventions for optimizing the use of antimicrobials to
achieve the best clinical outcomes while minimizing adverse effects, reducing resistance to antibiotics, and decreasing
unnecessary costs.
\item AMR prediction: forecasting the evolution and prevalence of resistance patterns, informing preventive measures and stewardship strategies.
\item New antimicrobial discovery: developing novel drugs to combat resistant pathogens, addressing gaps in current treatment options.
\end{itemize}
In the inner circle in Figure~\ref{fig:introduction_overview} (in purple), we illustrate the connection between these AMR tasks.
Specifically, AMR surveillance acts as the foundation.
Building upon surveillance data, AMR prediction can effectively inform preventive and treatment measures for antimicrobial stewardship, and address the gaps in current treatment options to facilitate new antimicrobial discovery.


Meanwhile, as we observed in recent AMR research, while ML-based analytical methods enhance quality and efficiency, they also face challenges from the increasing complexities of datasets and models. As illustrated in the outer circle of Figure \ref{fig:introduction_overview} (in green), it is essential to discuss the impacts of data noise and bias on AMR research and to outline strategies for their mitigation during data collection, handling, and model development, ensuring the reliability and accuracy of machine learning applications in this crucial field.

In brief, this comprehensive survey navigates the complex landscape of AMR, exploring not only the basics of how antibiotics work and how bacteria resist but also the evolving methods used to combat AMR. With a focus on data-driven approaches, including AI models, we deeply study the interconnected challenges of surveillance, prediction, new antimicrobial discovery, stewardship, and driver analysis,
emphasizing the report and analysis of data challenges and their mitigation methods in diverse tasks of the AMR domain. The paper aims to illuminate how data science, especially machine learning, plays a crucial role in understanding, addressing, and tackling the multifaceted challenges posed by AMR.
Specifically, we provide standards, potential risks, and solutions for privacy preservation from the perspective of data science to address relative issues in the AMR domain, as this healthcare field highly involves sensitive information such as personal identification information.

The rest of this paper is organized as follows. Section 2 describes AMR tasks in detail. 
This section offers insights into how data-driven approaches are being integrated into these tasks to enhance performance and innovation. Section 3 focuses on data collection to serve these tasks in the AMR domain, providing an overview of the primary sources and methods used in gathering data for AMR research, as well as challenges in data use.  Section 4 addresses mitigating data challenges in these tasks, discussing the issues of data noise and biases specific to the AMR domain. It elaborates on strategies to mitigate these challenges during critical phases of data preparation and model development, thereby enhancing the reliability and effectiveness of ML applications in AMR research.
As privacy-preserving techniques are often involved in handling clinical data, this section ends with a discussion about the side effects of privacy-preserving techniques in AMR research.
This structured survey provides an understanding from a data-driven perspective of both the potential and the challenges of applying machine learning in the fight against antimicrobial resistance.

\section{Data Analytics Tasks in AMR Research}


This section describes specific AMR tasks, including prediction, new antimicrobial discovery, stewardship, and driver analysis. By examining the latest techniques and tools, we particularly focus on how machine learning is utilized to tackle the complex challenges associated with these tasks, highlighting the objectives and analytical methods. 
Table \ref{table:AMR_Analysis} provides a summary of these tasks. 

\begin{table}[th]
    \small
    \centering
    \begin{tabular}{|>{\centering\arraybackslash}p{2.1cm}|p{3.2cm}|p{5.0cm}|p{4.8cm}|}
        \hline
        \textbf{AMR Tasks} & \textbf{Task Description} & \textbf{Data Used} & \textbf{Commonly-used Analytical Methods} \\
        \hline
        \textbf{AMR Prediction} & 
        Predicting the presence and spread of antimicrobial resistance genes (ARGs) in clinical, environmental, or genomic datasets. &
        \begin{itemize}[leftmargin=*, noitemsep, topsep=0pt]
            \item Genomic sequence data~\cite{Ren2022, nguyen2019using}
            \item Antibiotic resistance genes (e.g., UNIPROT, CARD, ARDB datasets)~\cite{arango2018deeparg}
            \item Mass spectra data~\cite{weis2022direct}
            \item \textit{Escherichia coli} antibiotic resistance knowledge graphs~\cite{Youn2022}
        \end{itemize} & 
        \begin{itemize}[leftmargin=*, noitemsep, topsep=0pt]
            \item Hidden Markov Models
            \item Machine learning (e.g., XGBoost, Random forest, SVM)
            \item Deep learning (e.g., CNNs)
            \item Link prediction using knowledge graphs constructed from public sources
        \end{itemize} \\
        \hline
        \textbf{Antimicrobial Stewardship} & 
        Supporting decision-making for prescribing antimicrobials and optimizing treatment plans to reduce AMR risks in healthcare settings. &
        \begin{itemize}[leftmargin=*, noitemsep, topsep=0pt]
            \item Prescription records, electronic health records (EHRs)~\cite{Kanjilal2020}
            \item Community-acquired infection data (e.g., UTIs data from Maccabi Healthcare Services)~\cite{yelin2019personal}
        \end{itemize} & 
        \begin{itemize}[leftmargin=*, noitemsep, topsep=0pt]
            \item Machine learning (e.g., Logistic regression, Gradient boosting decision trees, Random forest)
            \item Deep neural networks
            \item Hybrid rule-based approaches
        \end{itemize} \\
        \hline
        \textbf{AMR Driver Analysis} & 
        Identifying key factors and behaviours that contribute to the spread of AMR, considering factors such as human-animal-environment interactions or socioeconomic and governance. &
        \begin{itemize}[leftmargin=*, noitemsep, topsep=0pt]
            \item AMR prevalence data (e.g., from the Global Burden of Disease)
            \item Antimicrobial usage data (e.g., the IQVIA MIDAS database)
            \item Socioeconomic and governance data (e.g., from the World Health Organization (WHO), the World Bank DataBank, and the Center for Disease Dynamics Economics and Policy)
            \item Human-animal-environment data (e.g., ResistanceMap)
        \end{itemize} & 
        \begin{itemize}[leftmargin=*, noitemsep, topsep=0pt]
            \item Multivariable logistic regression
            \item Bayesian networks
            \item Compartmental models (e.g., ordinary differential equations)
            \item Systems mapping and counterfactual analysis
        \end{itemize} \\
        \hline
        \textbf{Novel Antimicrobial Discovery} & 
        Utilizing bioinformatics and computational models to identify and design new antimicrobial agents, including novel classes of antibiotics or natural compounds. &
        \begin{itemize}[leftmargin=*, noitemsep, topsep=0pt]
            \item Molecular structures (e.g., Drug library from the US Food and Drug Administration (FDA), Natural compounds from Human \& Original screened dataset)~\cite{wong2023discovery}
            \item Antimicrobial peptides database~\cite{li2022amplify}
            \item Metagenomic data~\cite{ma2022identification}
        \end{itemize} & 
        \begin{itemize}[leftmargin=*, noitemsep, topsep=0pt]
            \item Deep learning (e.g., Attention mechanisms)
            \item Natural language processing (e.g., Directed-message passing)
            \item Explainable graph neural networks
            \item Attribute-controlled generative models and molecular dynamics simulations
        \end{itemize} \\
        \hline
    \end{tabular}
    \caption{A summary of AMR tasks.}
    \label{table:AMR_Analysis}
\end{table}

\subsection{AMR Prediction} 

With the alarming rise in antibiotic resistance, it is particularly important to select optimal antibiotic treatments, as the random use of broad-spectrum antibiotics will further enhance the resistance compared to the targeted use of narrow-spectrum antibiotics. AMR prediction can be used to optimise the prescription of antibiotics. For example, \citet{weis2022direct} applied the machine learning method to clinical mass spectra data, enabling efficient and low-cost microbial identification. Their method reduces the time of AMR diagnostics compared to the conventional culture-based method, enabling  precise antibiotic prescriptions.

The predominant data type used in AMR prediction centers around the genomic sequence data, with a particular emphasis on antibiotic resistance genes (ARGs). \citet{arango2018deeparg} showcase the use of manually curated ARG databases, combining from public sources like the UniProt (Universal Protein Resource Database)~\cite{apweiler2004uniprot}, CARD (Comprehensive Antibiotic Resistance Database)~\cite{jia2016card} and ARDB (Antibiotic Resistance Genes Database)~\cite{liu2009ardb}. 
\citet{weis2022direct} introduce an additional way by leveraging mass spectra data from clinical isolates, offering real-world clinical insights into microbial identification and antimicrobial resistance prediction. This study introduces a new dimension for comprehensive understanding of the genetic basis of resistance.
\citet{Youn2022} introduce an innovative approach by incorporating knowledge graphs that integrate information from diverse sources, such as antibiotic resistance information, gene-regulatory relations, and biological impacts. By utilizing these knowledge graphs, their approach provides a more comprehensive understanding of the complex factors influencing antibiotic resistance.

AMR predictions are in a transition from traditional statistical method based to more complex machine learning based methods in order to accommodate more observable features. For example, \citet{gibson2015improved} applied the traditional statistical method, hidden Markov models in their study to identify ARG. Afterwards, \citet{nguyen2019using} showcase the application of  extreme gradient boosting-based regression methods for predicting antimicrobial Minimum Inhibitory Concentrations (MICs, where a lower MIC indicates a better anti-bacterial effect) using genomic data. \citet{Ren2022}  give multiple machine learning models for AMR prediction, including linear regression, support vector machines (SVM), random forests, and convolutional neural networks (CNNs). Other studies also demonstrated superior prediction performance with deep learning approaches \cite{arango2018deeparg,Youn2022,li2021hmd}. Complex machine learning methods are often used to deal with complex data such as genomic sequences to extract hidden patterns. 

\subsection{Antimicrobial Stewardship}

While AMR prediction focuses on predicting the likelihood of a specific microorganism being resistant to a particular antibiotic, antimicrobial stewardship refers to a collection of coordinated interventions designed to optimise the use of antimicrobials to achieve the best clinical outcomes. It aims to minimize adverse effects, reduce antibiotic resistance, and reduce the cost. These interventions aim to optimise antimicrobial drug regimens by ensuring the selection of the most appropriate drug, dose, duration, and route of administration. In this context, the use of machine learning techniques has gained increasing importance in enhancing antimicrobial stewardship efforts.

The primary data sources used by these machine learning methods are electronic health records (EHRs). EHRs provide comprehensive and detailed patient information, which is essential for developing accurate predictive models. Several studies have leveraged EHR data to build and test their models \cite{Kanjilal2020,lee2022hybrid,beaudoin2016evaluation,oonsivilai2018using}. Additionally, other data sources have also been utilized. For instance, \citet{yelin2019personal} used data from Maccabi Healthcare Services, which included community- and retirement home-acquired urinary tract infections (UTIs).
Antimicrobial stewardship can have different focuses, for example, \citet{beaudoin2016evaluation} targeted the prediction of inappropriate prescriptions of piperacillin–tazobactam. They applied a temporal induction of classification models for the clinical decision support system and this method allowed for identifying patterns in prescription data over time.
Building on another objective of reducing prescription errors, \citet{lee2022hybrid} utilized a hybrid method combining a rule-based approach with an advanced deep neural network for robust and accurate prescription error prediction.
In another study, \citet{yelin2019personal} aimed to predict mismatched treatments, defined as instances when the sample is resistant to the prescribed antibiotic, by leveraging logistic regression, decision trees, and gradient-boosting decision trees on personal clinical history data.
\citet{oonsivilai2018using} expanded the scope to include predicting susceptibility to antibiotics by applying a comprehensive set of machine learning models, including logistic regression, decision trees, random forests, GBDTs, support vector machines (SVMs), and K-nearest neighbours (KNNs) to guide empiric antibiotic prescribing.
Extending the focus on prescription accuracy, \citet{Kanjilal2020} focused on the proportion of recommendations not only for inappropriate antibiotic therapies but also for second-line antibiotics. They employed logistic regression, decision trees, and random forest models to develop a decision algorithm for outpatient antimicrobial stewardship in uncomplicated UTIs.

With the growing data collected, machine-learning techniques have shown great potential in improving prescription practices and reducing antibiotic resistance in antimicrobial stewardship. 

\subsection{AMR Driver Quantification}

Understanding the drivers of antimicrobial resistance (AMR) in a specific population or system is significant in combating its spread and mitigating its impact on public health. Driver analysis in the AMR domain aims to identify and quantify the factors that contribute to the emergence and dissemination of antimicrobial-resistant pathogens. 

Analyzing antimicrobial resistance (AMR) drivers from the perspective of transmission across human, animal, and environmental domains reveals critical insights into the association of factors shaping resistance patterns. Studies such as \cite{ALLEL2023e291,booton2021one} investigate the One Health framework, quantifying the relative impacts of human, animal, and environmental use and transmission of antibiotics. \citet{ALLEL2023e291} reveals associations between animal antimicrobial consumption and AMR in food-producing animals, while \citet{booton2021one} identifies that human antibacterial usage is the primary driver in human antibacterial resistance (ABR). Furthermore, research by \citet{xie2018antibiotics} explains the pathways through which antibiotics and antibiotic resistance disseminate from animal manures to the soil. These investigations emphasize the complex dynamics of the human-animal-environment interaction and also highlight the need for holistic strategies to address AMR effectively. Additionally, the Antimicrobial Resistance Systems Map presented by \citet{ukamrmap2014} provides a comprehensive overview of the interconnected elements influencing AMR development, including factors related to food-producing animals, the environment, healthcare facilities, community, pharmaceuticals, and vaccinations.

From a socioeconomic perspective, factors such as governance, education, economic indicators, and healthcare infrastructure play crucial roles in AMR prevalence. Studies such as \cite{Collignon2018,maugeri2023socio} examine the anthropological and socioeconomic determinants contributing to global AMR prevalence, emphasizing the importance of addressing broader societal factors beyond antibiotic consumption alone. These investigations highlight the need for multifaceted interventions that involve governance reforms, improved access to healthcare, and enhanced sanitation practices to mitigate AMR effectively. Moreover, research by \citet{AWASTHI2022133} integrates AMR data with the global burden of disease (GBD), governance (WGI), and finance data sets in an attempt to find AMR’s unbiased and actionable determinants. Additionally, the study by \citet{vikesland2019differential} explains the importance of considering the development status of low to middle-income countries (LMICs) and high-income countries (HICs) as a significant factor in global AMR dynamics, emphasizing the diverse physical, social, and economic circumstances within LMICs that potentially benefit AMR dissemination.

Beyond human-animal-environment transmission and socioeconomic factors, other angles exist to analyze AMR drivers in understanding the complexity of resistance dynamics. For instance, \cite{CHEN2008639} investigates the role of bacteremia in previously hospitalized patients and identifies prolonged effects from previous hospitalization as well as risk factors for antimicrobial-resistant bacterial infections. This research emphasizes the importance of healthcare-associated factors in driving AMR, highlighting the need for enhanced infection control measures and antimicrobial management practices in healthcare institutions.

Additionally, the data utilized across these studies are drawn from diverse sources, including global organizations and specialized databases. 
On a global level, \citet{ALLEL2023e291} draw upon data provided by the World Health Organization (WHO), the World Bank DataBank, and the Center for Disease Dynamics Economics and Policy, emphasizing the significance of large-scale institutional data for understanding global trends in AMR drivers. Similarly, \citet{Collignon2018} also leverage the WHO report and the World Bank, alongside antibiotic consumption data from the IQVIA MIDAS database and AMR data from ResistanceMap. Also, \citet{maugeri2023socio} utilize the World Bank, as well as a multitude of datasets, including community consumption of antibiotics from the ESAC-Net database, pathogen data from The European Centre for Disease Prevention and Control (ECDC) atlas, and political rights scores from The Freedom House and pathogen data from The European Centre for Disease Prevention and Control (ECDC) atlas.
Moreover, research by \citet{AWASTHI2022133} integrates AMR data with the Global Burden of Disease (GBD), Governance (WGI), and finance data sets for unbiased analysis. 
On a national level, \citet{booton2021one} quantifies from Thailand's AMR prevalence data for human, animal, and environmental sectors. And \cite{xie2018antibiotics} reviews publications and data concerning veterinary antibiotic usage, and its impact on soil resistance dynamics in China, highlighting the transmission of resistance from animal manures to environmental reservoirs.
On a more localized level, research by \cite{CHEN2008639} involves data collected from a cohort of 789 patients enrolled in a year-long post-hospitalization study, providing insights into the drivers of AMR within a healthcare facility. 
These varied data sources enrich the analysis by providing a holistic view of the factors influencing antimicrobial resistance at both global and local levels.

Among these studies, various analytical methods have been employed to clarify the contributions of different drivers to AMR. Multivariable logistic regression is one of the most popular methods which have been used in \cite{ CHEN2008639,ALLEL2023e291,Collignon2018,maugeri2023socio}. But \citet{booton2021one} builds a compartmental model using ordinary differential equations to describe the relationship between resistant bacteria in the three compartments: humans, animals and the environment. Additionally, in \cite{ukamrmap2014}, systems mapping techniques are utilized to conceptually represent the complex interaction between different elements influencing AMR development. Furthermore, \citet{AWASTHI2022133} applies Bayesian networks, counterfactual analysis, and supervised machine learning algorithms to uncover AMR determinants.

The findings from these studies uncover the multifactorial nature of AMR, with human antibacterial usage often identified as a primary driver \cite{booton2021one}. However, the reduction of antibiotic consumption alone may not suffice to mitigate AMR, as the transmission of resistance genes and other socio-environmental factors also play crucial roles \cite{Collignon2018}. 
This finding is also observed by \citet{vikesland2019differential}, noting that LMICs exhibit lower average antibiotic consumption yet experience more severe AMR.
Governance emerges as a significant contributing factor, highlighting the importance of effective policies and regulations in combating AMR \cite{maugeri2023socio}. The integration of diverse datasets, analytical techniques, and interdisciplinary approaches is essential in quantifying AMR drivers comprehensively and devising effective strategies for mitigating the global threat posed by antimicrobial resistance.

\subsection{Novel Antimicrobial Discovery}

In the ongoing challenge against AMR, the quest for new antimicrobial drugs remains a paramount concern, especially with recent strides in deep learning methods showcasing remarkable performance, illuminating promising new paths of inquiry. In 2017, Wright et al. \cite{wright2017opportunities} provide a prior insight into antibiotics exploration status, highlighting potential solutions such as antibiotic adjuvants, alternatives of antibiotics such as antivirulence compounds and biofilm inhibitors, and techniques of synthetic biology. Until 2020, Stokes et al. \cite{stokes2020deep} advanced antibiotic discovery by fully leveraging AI for the first time, representing a notable step forward in the field. Their directed-message passing graph neural network model, trained on diverse molecular datasets, exhibits promising results in predicting antibiotics based on molecular structures, with one candidate demonstrating broad-spectrum antibiotic activities. Then, Das et al. \cite{das2021accelerated} push the frontier with accelerated antimicrobial discovery frameworks, merging attribute-controlled deep generative models and molecular dynamics simulations to expedite the identification of antimicrobial candidates. Li et al. \cite{li2022amplify} harness the power of attention mechanisms in AMPlify, a deep learning model tailored for the discovery of antimicrobial peptides effective against WHO-priority pathogens. Furthermore, Ma et al. \cite{ma2022identification} delve into metagenomic data to uncover antimicrobial peptides within the human gut microbiome, employing a series of natural language processing neural network models. More Recently, Wong et al. \cite{wong2023discovery} innovatively introduced an explainable graph neural network to discover new structural classes of antibiotics by leveraging large chemical libraries.

Meanwhile, with the advances in methods, particularly the integration of deep learning networks, the diversity and complexity of data available in this field have expanded significantly. This encompasses a wide range of molecular structures, including those from drug libraries, natural products, various antibiotics, and other compounds \cite{stokes2020deep,wong2023discovery}, as well as antimicrobial peptides \cite{das2021accelerated,li2022amplify} and metagenomic data \cite{ma2022identification}.

In sum, the feature of employed data analytical methods varies across different AMR tasks, ranging from traditional statistical analysis to deep neural networks.
AMR surveillance mainly focuses on data collection and extracting general descriptional information from data, and therefore, works in this task mainly utilize basic statistical methods, for example, basic statistics (e.g., data distribution and average), and statistical tests (e.g., t-test).
AMR driver quantification and antimicrobial stewardship extract more information from AMR-related data, and therefore, they involve more advanced methods, like machine learning.
Among the aforementioned AMR tasks, AMR prediction and novel antimicrobial discovery, especially when tackling genetic data, employ the most complex data analytical models, including advanced deep learning models.


\section{AMR Data Sources, Collection, and Challenges}
The previous section mentioned specific examples of datasets used in AMR tasks. Here, we provide a broader overview of the overall landscape of data in this field. 
Data used in AMR research are commonly sourced from diverse origins. Given the varied focuses across different research domains of AMR, a range of data types are employed, as illustrated below.

\begin{itemize}
\item AMR prediction. Genomic data plays a key role in identifying and predicting antimicrobial resistance genes (ARGs)~\cite{arango2018deeparg, nguyen2019using, Ren2022}. Given the time-consuming and low-throughput nature of traditional antimicrobial susceptibility testing (AST), which applies only to cultivable bacteria, whole-genome sequencing (WGS) has become a routine method for ARG profiling. This involves comparing genomic sequences with databases of known ARGs, such as CARD, ResFinder, and UniProt. These databases are manually curated, providing molecular sequence references for predicting AMR genotypes from genomic data. To enhance ARG profiles, databases from different sources are often integrated~\cite{Chiu2019, arango2018deeparg}. Many studies demonstrate the potential of machine learning methods for predicting AMR by combining sequencing approaches, well-known databases, and phenotypic information~\cite{Ren2022}. Additionally, matrix-assisted laser desorption ionization-time of flight (MALDI-TOF) mass spectrometry, commonly used for microbial species identification, has also been applied to AMR prediction~\cite{weis2022direct}.

\item Antimicrobial stewardship. As discussed, the primary goal of antimicrobial stewardship is to optimise antimicrobial use and aid in the selection of suitable treatment regimens~\cite{DYAR2017793}. Traditionally, clinicians have relied on patient clinical data as their main decision-making source. Despite the existence of computerised decision systems like TREAT for antibiotic treatment~\cite{Leibovici2013}, their adoption has been limited due to their specialised nature~\cite{Anahtar2021}. The emergence of electronic health records (EHRs) enables the integration of data-driven approaches to enhance clinician decisions. For instance, Kanjilal et al.~\cite{Kanjilal2020} employ machine learning on EHR data to predict antibiotic susceptibility and develop a decision algorithm recommending the narrowest possible antibiotic to which a specimen is susceptible. EHRs include patient-relevant data, such as demographics, clinical information, pharmacy records, and laboratory results.

\item AMR driver analysis. Quantifying the drivers of AMR in humans requires data from diverse domains and sources. Chatterjee et al.\cite{chatterjee2018quantifying} conducted a systematic review of AMR studies published between January 1, 2005, and February 14, 2018, identifying 88 drivers across 5 key domains: patient clinical history (e.g., underlying disease), demographics (e.g., age and ethnicity), healthcare factors (e.g., invasive procedures), antibiotic usage (e.g., prior antibiotic exposure), and community-level influences (e.g., water and animals). Expanding this perspective, Collignon et al.\cite{Collignon2018} analysed AMR in relation to global antibiotic consumption, incorporating anthropological and socioeconomic factors. Their work highlights the role of governance, education, GDP per capita, healthcare expenditure, and community infrastructure, using data sourced from the World Bank’s DataBank. Building upon this work, Maugeri et al.~\cite{maugeri2023socio} explored the influence of demographic and freedom-related factors on antibiotic consumption and AMR across 30 European countries, offering additional insights into regional variations.

\item Novel antimicrobial discovery. The search for new antibiotics often involves the screening of large chemical libraries, which can contain hundreds of thousands to a few million molecules. Machine learning approaches provide an efficient and cost-effective means to explore these vast chemical spaces in silico~\cite{stokes2020deep, wong2023discovery, li2022amplify}. For instance, in~\cite{stokes2020deep}, researchers have utilised multiple chemical libraries to predict antibiotic activity, leading to the discovery of a molecule from the Drug Repurposing Hub that is structurally divergent from known antibiotics. The Drug Repurposing Hub is a curated and annotated repository that includes FDA-approved drugs, clinical trial drugs, and pre-clinical tool compounds. It provides detailed information on their chemical structures, clinical trial status, mechanism of action and protein targets. 
\end{itemize}

Table~\ref{table:AMR_data_sources} summarises the types of data and publicly accessible sources used for different AMR tasks. As shown, certain data types are shared across multiple tasks. For instance, antimicrobial resistance data are utilised in both antimicrobial stewardship and AMR driver analysis. To avoid redundancy, the column listing example public data sources includes mainly those that correspond to data types unique to each task. It is worth noting that publicly available electronic health record data remain scarce due to privacy concerns. The data collection process underlying these tasks will be detailed in the following section.
\begin{table}[t]
    \small
    \centering
    \begin{tabular}{|p{2.5cm}|p{8cm}|p{3.5cm}|}
        \hline
        \textbf{AMR Tasks} & \textbf{Common Data Types} & \textbf{Example Public Data Sources}\\
        \hline
        \textbf{AMR Prediction} & 
        \begin{itemize}[leftmargin=*, noitemsep, topsep=0pt]
            \item Genome sequence data
			\item MALDI-TOF mass spectra data and resistance profile
			\item Phenotypic resistance profiles
                \item ARG databases
        \end{itemize} & 
        \begin{itemize}[leftmargin=*, noitemsep, topsep=0pt]
           \item UniProt\tablefootnote{\url{ https://www.uniprot.org/}.}
           \item CARD\tablefootnote{\url{ https://card.mcmaster.ca/}.}
           \item ResFinder\tablefootnote{\url{http://genepi.food.dtu.dk/resfinder}.}
           \item DRIMAS\tablefootnote{\url{https://datadryad.org/stash/dataset/doi:10.5061/dryad.bzkh1899q}.}
           \end{itemize}\\
        \hline
        \textbf{Antimicrobial Stewardship} & 
        \begin{itemize}[leftmargin=*, noitemsep, topsep=0pt]
        	\item Patient information (Demographics, infection origin, clinical history, specimen type)
			\item Antimicrobial resistance (AMR) data
			\item Antimicrobial consumption (AMC) and usage (AMU) data
			\item Phenotypic resistance profiles
        \end{itemize} &
        \begin{itemize}[leftmargin=*, noitemsep, topsep=0pt]
            \item PhysioNet\tablefootnote{\url{https://physionet.org/}.}
        \end{itemize}\\
        \hline
        \textbf{AMR Driver Analysis} & 
        \begin{itemize}[leftmargin=*, noitemsep, topsep=0pt]
            \item Antimicrobial resistance (AMR) data
			\item Antimicrobial consumption (AMC) data 
			\item Data on transmission pathways 
			\item Socioeconomic and environmental data
        \end{itemize} & 
        \begin{itemize}[leftmargin=*, noitemsep, topsep=0pt]
            \item ResistanceMap\tablefootnote{\url{https://resistancemap.onehealthtrust.org/}.}
            \item GLASS\tablefootnote{\url{https://www.who.int/data/gho/data/themes/topics/global-antimicrobial-resistance-surveillance-system-glass}.}
            \item ECDC\tablefootnote{\url{ https://www.ecdc.europa.eu/en}.}
            \item DataBank\tablefootnote{\url{https://databank.worldbank.org/}.}
        \end{itemize}\\
        \hline
        \textbf{Antibiotics Discovery} & 
        \begin{itemize}[leftmargin=*, noitemsep, topsep=0pt]
            \item Molecular data (both natural products and antibiotics) 
            \item Sequence data of antibiotics and resistance gene 
		\item Resistance profiles of antibiotics
        \end{itemize} & 
        \begin{itemize}[leftmargin=*, noitemsep, topsep=0pt]
        \item ZINC-20\tablefootnote{\url{https://zinc20.docking.org/}.}
        \item The Drug Repurposing Hub\tablefootnote{\url{https://www.broadinstitute.org/drug-repurposing-hub}.}
        \item DADP\tablefootnote{\url{http://split4.pmfst.hr/dadp/}.}
        \end{itemize}\\
        \hline
    \end{tabular}
    \caption{Common data types and sources for AMR data analytics tasks.}
    \label{table:AMR_data_sources}
\end{table}

\subsection{AMR Data Collection and Surveillance}

AMR-related data are collected through routine, periodic, or sporadic activities, similar to other data collection practices. Routine data collection activities, exemplified by medical records, are conducted on a regular, continuous basis at service delivery points such as health facilities, pharmacies, or laboratories. Conversely, non-routine activities such as surveys and interviews are conducted periodically or as one-time efforts to answer specific questions. While surveys may be costly and difficult to repeat, they often yield higher-quality data that are more representative of the target population due to their focused nature.

An example of routine data collection is \textit{surveillance}, which the World Health Organization (WHO) defines as “the continuous, systematic collection, analysis, and interpretation of health-related data needed for action”\footnote{\url{https://www.who.int/westernpacific/emergencies/surveillance}.}. AMR surveillance aims to detect and monitor changes and trends in microbial populations, including drug-resistant organisms and resistance determinants like genes and mechanisms\footnote{\url{https://www.fao.org/antimicrobial-resistance/key-sectors/surveillance-and-monitoring/en}.}. Effective AMR surveillance systems should capture patients' demographic and clinical profiles while integrating with other monitoring platforms, such as those tracking antimicrobial consumption (AMC) and the quality and supply chains of antimicrobial drugs. For instance, the Global Antimicrobial Resistance and Use Surveillance System (GLASS)~\cite{world2023glass, ajulo2024global}, established by WHO through the May 2015 World Health Assembly action plan on antimicrobial resistance\cite{10665-193736}, provides a standardized framework for collecting, analyzing, and sharing antimicrobial resistance (AMR) data from participating countries. GLASS collaborates with national and regional AMR surveillance networks, such as the Australian Commission on Safety and Quality in Health Care\footnote{\url{https://www.safetyandquality.gov.au/our-work/antimicrobial-resistance}.}.

As part of a One Health approach, human AMR surveillance data should be combined with data from animal, agricultural, and environmental sectors to provide a holistic view of AMR~\cite{FROST2021222, world2014antimicrobial}. However, AMR data collection in humans remains more comprehensive and systematic than in other sectors (e.g., animals, plants, and the environment) due to the presence of well-established health facilities in many parts of the world. To facilitate systematic collection and analysis, international and national health organisations, such as WHO, have published standard AMR surveillance manuals. These manuals guide healthcare and medical practitioners in collecting and analysing data at various levels, such as hospital and community levels. They also standardise key elements, including the categorization of microbes and antimicrobials, antimicrobial consumption doses, and methodologies for data collection, such as reporting frequency and contextual details about hospitals and patients. Under these guidelines, organisations regularly report on international and national human AMR status, as exemplified in publications like~\cite{aura2023, world2014antimicrobial}, based on indicators derived from standardised data.

Beyond humans, AMR-related data are also collected in other sectors. Examples include studies on antimicrobial usage in domesticated animals~\cite{Hur2020}, antibiotic-resistant lineages in Australian silver gulls~\cite{wyrsch2022urban}, AMR features such as antibiotics, antibiotic-resistant bacteria (ARB), and antibiotic-resistance genes (ARGs) in water sources (e.g., wastewater, recycled water, or surface water)~\cite{liguori2022antimicrobial, wang2018occurrence, blaak2015multidrug}, ARG types and concentrations in manure and soil from swine farms~\cite{zhu2013diverse}, and antibiotic usage in water and soil~\cite{zhang2015comprehensive}.

However, compared to AMR data collection in humans, data collection in these sectors is less systematic, often fragmented, and hindered by challenges such as the lack of organised monitoring systems and reliance on individual research teams. Realising the importance of the more comprehensive One Health approach, national and international organizations have started advancing AMR surveillance across sectors. For instance, WHO published a manual for One Health surveillance to monitor an indicator—the extended spectrum beta-lactamase (ESBL)-producing Escherichia coli—across human, animal, and environmental sectors~\cite{world2021onehealth}.

\subsection{Challenges with Data Collection and Use}

The collection and use of AMR data present many challenges. Firstly, a significant challenge lies in the unavailability of data, particularly pronounced in regions such as low- and middle-income countries (LMICs) where robust surveillance systems for AMR are lacking. This absence leads to significant gaps in understanding the true extent of antimicrobial resistance. Factors such as weak laboratory capacities, inadequate governance of health systems, deficient health information systems, and limited resources contribute to the challenges faced by LMICs in establishing effective surveillance mechanisms~\cite{iskandar2021surveillance}. Even when data are available, they often suffer from limited coverage and lack crucial details. In a study on global trends of hospital-associated infections due to hospital-acquired resistant infections (HARI)~\cite{balasubramanian2023global}, despite covering over 90\% of the global population, findings were reported from only 99 countries, and the data reported may not be nationally representative or instead biased towards larger tertiary care hospitals. Moreover, only 11 low-income countries report hospital-based point prevalence surveys, often based on small samples.  Even if data is available in a particular location, reports on hospitalisation rates and resistance for individual drug–pathogen combinations are scarce, thus introducing uncertainties when extrapolating country-level infection rates and resistance prevalence from the hospital-level data. Similar challenges are faced by other global or regional studies on AMR~\cite{Collignon2018,maugeri2023socio}. 

Additionally, data from different sources often exhibit heterogeneity in terms of structure, format, and content. Data might be structured (e.g., EHRs), unstructured (e.g., surveillance reports), or fall in between. Even within structured data, there are variations in data representations. For example, diagnoses may be captured in unstructured clinical notes, or structured data such as Internation Classification of Diseases (ICD) codes, while laboratory results may be labelled with specific or nonspecific Logical Observation Identifiers Names and Codes (LOINC) codes~\cite{bastarache2022developing}. Additionally, there are divergences in data collection methods, diagnostic criteria, and reporting practices, contributing to challenges in data use and undermining the reliability of analysis results. 

Furthermore, data integrity is frequently compromised by various quality issues Errors introduced during data entry or integration pose significant concerns. For instance, manually curated gene databases often contain inaccuracies such as sequence annotation typos and classification errors, which propagate when integrating data from multiple sources without validation (~\cite{Chiu2019, arango2018deeparg}). Nonrandom missing values in the data can skew results and impact the generalisability of findings (~\cite{Paxton2013}). Insufficient metadata accompanying the data complicates interpretation and use, as critical contextual information may be absent. Conflicting facts from disparate sources may lead to inconsistencies when integrating data, undermining the reliability of analytical outcomes. Finally, redundancies or duplicates or imbalanced distributions in the data can introduce noise and bias if not addressed.

\section{Mitigating Data Challenges in ML-based AMR Research}

\begin{figure*}[htbp]
 \centering
  \includegraphics[width=1.0\textwidth]{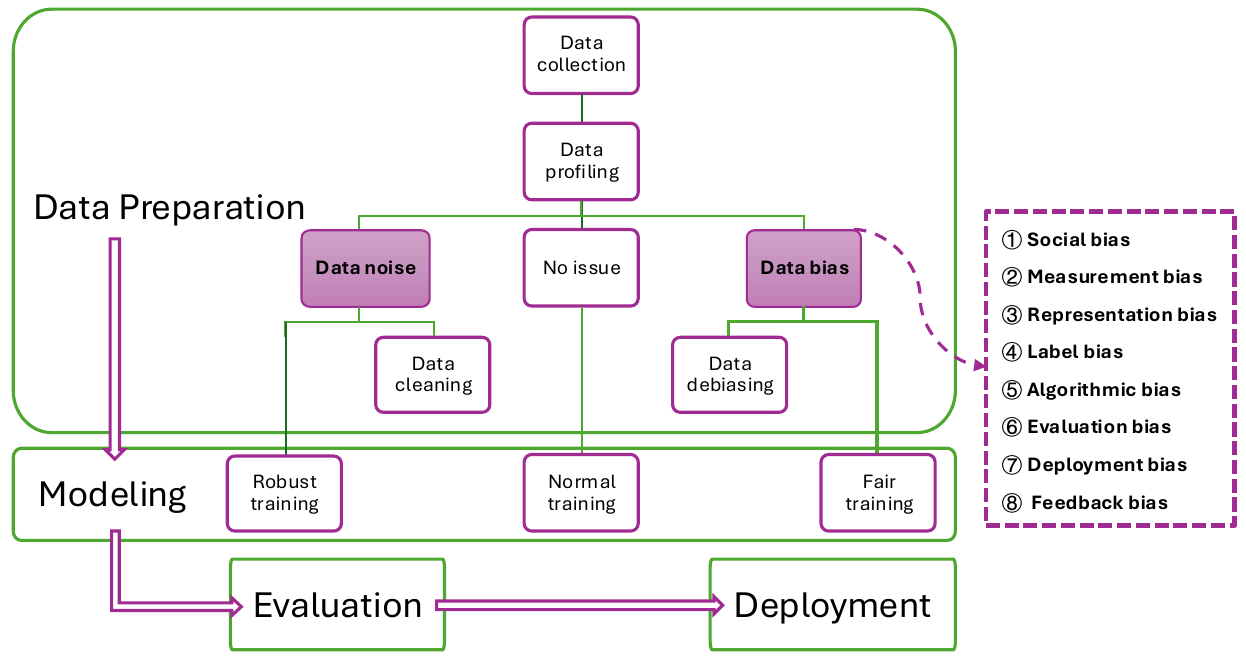}
  \caption{Addressing data issues, mainly noise and bias, spans multiple phases of the machine, notably during the data handling and modelling phases}
  \label{fig:pipeline}
\end{figure*}

As we observed in Section 2, AMR research is increasingly adopting machine learning (ML) and deep learning (DL) methods to tackle complex tasks. However, these methods come with inherent data challenges, many of which were outlined in Section 3.2. For example, heterogeneity in data formats and collection methods can often lead to noise. Similarly, integrity issues like missing metadata, incomplete datasets, and skewed data collection practices can contribute to bias, such as representation bias or measurement bias. These challenges, while specific to AMR contexts, also align with general ML workflows.
Figure \ref{fig:pipeline} shows the mitigation process for data challenges, primarily including data noise and bias, spans various phases of the general ML workflow, particularly during data handling and model development phases. 

For the main AMR tasks, the initial step typically involves data collection, a process sensitive to various forms of noise and bias. Data noise refers to irrelevant or erroneous data that can obscure the underlying patterns and relationships within the dataset, while data bias pertains to systematic errors or inaccuracies that skew the dataset towards certain outcomes or perspectives. Addressing these issues is crucial for ensuring the reliability and accuracy of subsequent analyses. In this section, we delve into strategies for mitigating data noise and bias in AMR research, exploring both the phases of data preparation, modelling and evaluation.

\subsection{Mitigation Process in Data Collection and Preparation}\label{sec:pre}
Among the data quality issues listed in Section \ref{sec:intro}, several, namely data noise, social bias, measurement bias, representation bias, and label bias, can be mitigated during the data collection and preparation process.
Aligning with balanced and consistent principles in the data collection process is one of the most efficient ways to obtain low noise and unbiased data analytical results.
However, defining proper principles for each situation and ensuring each data collection site complies with the predefined principles are challenging, especially for large-scale data collection schemes.
We take the principles of WHO's GLASS \cite{world2023glass} for AMR surveillance, a major systematic AMR data collection approach, as an example to demonstrate the mitigation for the above data quality issues.

\citet{world2023glass} provide principles for nation-level members of the GLASS surveillance system, suggesting that each country gradually improves their data collection processing to achieve data denoise and debias goals by the following points.
\begin{enumerate}
\item Clearly define the AMR surveillance objectives so that the surveillance system can set up efficient data collection plans.
For instance, with a clear surveillance target, e.g., to assess human-centred or one health level AMR and to monitor hospital-level or wider health care facility level AMR, the surveillance system is able to build a comprehensive data collection roadmap to achieve the corresponding goals.
\item Include comprehensive populations in the surveillance system and set up a sufficient representation sampling size for each population to ensure the data sampling is geographically and demographically balanced.
Within each population, healthcare facilities of the target level should be gradually added to the surveillance system to ensure a balanced data sampling scheme.
\item Probability sampling methods can be used as guidance to design the surveillance system to ensure sufficient randomness and balance in data collection.
\item Regularly train surveillance sites for systematically identifying patients with suspected infection, presenting results of consistent departments and wards, and conducting consistent and rigorous laboratory testings.
When clinical quality-assured laboratories are absent, set quality-assured laboratories in strategic geolocations to maximize diagnostic coverage of healthcare facilities and deploy rapid sample referral from surveillance sites.
Promote routine communication between relevant teams, e.g., clinicians, laboratory personnel, the infection prevention and control team, and epidemiologists.
\end{enumerate}

Potentially, approaches (1), (2), and (3) aim at mitigating social bias, measurement bias, representation bias, and label bias. (4) mainly aims at reducing data noise while also mitigating measurement bias and label bias.
However, it is challenging to guarantee the principles at all sites nationwide, especially in developing and remote areas, and GLASS suggests gradually improving the surveillance system.

Data denoising is often used in the data preparation phase. Data denoising aims to remove or minimize noise from the dataset, thereby enhancing its quality and utility for subsequent analysis. The related studies mainly adopt strategies such as data validation, cleaning, and sanitization processes \cite{whang2023data}. For instance, \citet{arango2018deeparg} addresses data noise through database merging with duplicate removal, AMR annotation validation, and manual correction of categorization errors. Similarly, \citet{li2021hmd} remove identical and duplicate sequences for data cleaning, while \citet{gibson2015improved} manually curate annotation to deal with incomplete annotations. With data pre-processing, \citet{ma2022identification} preprocess their data by cross-checking and screening, and \citet{wang2018occurrence} preprocess raw data from hospital wastewater samples through trimming, denoising, and clustering. \citet{Chiu2019} perform validation and outlier detection to sanitize the data and rectify inaccuracies from obsolete annotations or redundant sequences and misclassified sequences. Furthermore, efforts to address missing data are evident in studies such as \cite{Hur2020}, where data with substantial missing values are excluded from analysis, and in \cite{Youn2022}, missing links are fixed by link prediction.

\subsection{Mitigation Strategies in Modeling}\label{sec:modeling}

In the preceding discussion, we primarily explored the collection, utilization, and analytical methods employed in AMR research across four core tasks: prediction, stewardship, antimicrobial discovery, and driver quantification.

An observation worth noting is the growing preference among researchers for machine learning (ML) and deep learning methods over traditional statistical approaches. For example, in the context of the AMR prediction task, this shift is underscored by several compelling reasons: 
\begin{enumerate}
\item \textbf{Complexity of Biological Data}: AMR involves intricate genetic variations and interactions among bacteria. Machine learning excels at capturing complex patterns within genomic data. The depth and adaptability of these models make them well-suited for identifying subtle genetic variations associated with antibiotic resistance genes (ARGs) \cite{arango2018deeparg}.
\item \textbf{Improved Predictive Accuracy}: Machine learning and deep learning models \cite{nguyen2019using,weis2022direct,Ren2022} have demonstrated superior predictive accuracy compared to traditional statistical methods. These models can handle the non-linear relationships inherent in genetic and clinical data, resulting in more accurate predictions.
\item \textbf{Adaptability to Diverse Data Types}: The shift towards machine learning corresponds with the incorporation of diverse data types beyond genomic data. For instance, mass spectra data from clinical isolates \cite{weis2022direct} and knowledge graphs \cite{Youn2022} contribute additional perspectives to the predictive models. Machine learning techniques can seamlessly integrate and analyze these multiple data dimensions, providing a more comprehensive understanding of AMR.
\item \textbf{Accelerated Predictions and Practical Applicability}: Machine learning models \cite{nguyen2019using,weis2022direct} have demonstrated good performance for accelerating the prediction process. This is particularly important in healthcare settings where timely and accurate predictions of antimicrobial resistance can influence treatment decisions.
\item \textbf{Flexibility and Generalization}: Machine learning models, being data-driven and flexible, can generalize well to new data. In contrast, traditional statistical methods may struggle to adapt to the complexities of evolving resistance patterns or new genetic information.
\end{enumerate} 




Robustness improvement in the modelling phase is crucial for developing machine learning models that can effectively handle noisy data and produce reliable predictions. 
Some methodologies include data imputation techniques employed by \cite{maugeri2023socio,AWASTHI2022133} to enhance the robustness of predictive models. Additionally, \citet{ma2022identification} enhances model robustness by combining outputs from multiple natural language processing models, thereby increasing the model's resilience to noisy input data. Furthermore, recent advancements in ML-based AMR research, particularly relating to deep learning approaches, have shown promise in improving model robustness and generalization capabilities, as presented by the latest bias-variance trade-off discussion \cite{belkin2019reconciling,yang2020rethinking}.
\subsubsection{Data Bias and Fair Modeling}

In the context of machine learning (ML) projects, biases can manifest at various stages, impacting the reliability and fairness of model outcomes. \citet{van2022overcoming} summarizes eight types of biases prevalent in ML projects, including social bias, representation bias, measurement bias, label bias, algorithmic bias, deployment bias, evaluation bias, and feedback bias. Social bias occurs when available data reflects existing biases in the relevant population prior to the creation of the ML model. Representation bias arises when the input data fails to adequately represent the relevant population. Measurement bias occurs when chosen features and labels are imperfect proxies for the true variables of interest. Label bias occurs when labelled data systematically deviates from the underlying truth categories. Algorithmic bias results from inappropriate technical considerations during modelling, leading to systemic deviations in outcomes. Deployment bias occurs when the ML model is used and interpreted in a different context than it was built for. Evaluation bias arises from the use of nonrepresentative testing populations or inappropriate performance metrics. 

In AMR research, representation bias and algorithmic bias are particularly prominent. To address representation bias during the data preparation phase, \citet{stokes2020deep} supplemented the training data with natural products to ensure a diverse chemical composition. \citet{vikesland2019differential} utilized randomized sampling and incorporated data from additional confounders. \citet{Ren2022} addressed severe data imbalance through down-sampling techniques. Additionally, \citet{arango2018deeparg} augmented the data using a low false-positive validation approach synthesized by randomly selecting partial ARG sequences. Similarly, data augmentation via synthetic minority oversampling \cite{AWASTHI2022133} and through knowledge inference over manually created rules \cite{Youn2022} were employed to mitigate representation bias.

Then, as previously discussed, despite efforts to mitigate data bias in the data preparation phase, residual bias may persist, necessitating further interventions during fair model training. \citet{das2021accelerated} address algorithmic bias through regularization techniques in the encoder. Additionally, \citet{wong2023discovery} advocates for the use of interpretable deep learning models to explain the decisions to enhance model transparency and mitigate algorithmic bias. Seeking interpretability is a way to address the data and model bias as it confirms the discovery based on human knowledge.

From the observation, biases prevalent in AMR research mainly revolve around representation and algorithmic biases, in addition, many works have used cross-validation approaches for mitigating potential evaluation bias \cite{ALLEL2023e291,li2022amplify,weis2022direct,li2021hmd,nguyen2019using,chandrasekaran2016chemogenomics}. The primary methods employed to mitigate data bias involve data debiasing techniques such as data augmentation and resampling, underscoring the importance of addressing bias at both the data preparation and modelling phases to ensure the reliability and fairness of ML-based AMR research outcomes.

\subsubsection{Improving Fairness in Other Medical Domains}

Transitioning from the discussion on bias mitigation in AMR research, we delve into the broader landscape of fairness improvement methods in other medical domains.

Healthcare often has disparities linked to societal biases such as geographic and financial issues and racial biases. When AI models are trained on unrepresentative databases or use improper proxies, it could lead to the unintentional amplification of existing biases, potentially resulting in systematic discrimination against specific groups, like racial discrimination.
\cite{obermeyer2019dissecting} discussed a widely used commercial prediction algorithm that exhibited racial bias, primarily attributed to inappropriate proxy (measurement bias) related to healthcare expenses rather than actual illness indicators, then reformulate the algorithm with proper proxy measurement to accurately identify and predict patients who need extra care.
\cite{pfohl2019creating} applied adversarial learning techniques to electronic health records, aiming to construct an inclusive atherosclerotic cardiovascular disease (ASCVD) risk prediction model. This model was specifically designed to ensure fairness across various gender and race groups, contributing to a more comprehensive and fair approach to healthcare analysis.
\cite{poulain2023improving} presented a strategy to address bias concerns by fostering collaboration among local healthcare institutions through a federated learning paradigm. They encouraged a fair federated learning model with sensitive information-free representation by incorporating adversarial debiasing and a fair aggregation method that is adaptable to diverse fairness metrics, particularly in the healthcare domain where electronic health records are employed.

For the specific discussion of biases, \cite{gianfrancesco2018potential} lists three main biases in machine learning algorithms using electronic health record data for diagnosis and treatment: (1) missing or incomplete data of certain patient populations which results in inaccurate predictions for these populations; (2) insufficient sample sizes which make data unrepresentative and lead to underestimation for certain patient populations; (3) misclassification or measurement errors which may be introduced by practitioners and make algorithms inaccurately learn and embed practitioner biases.
There are a series of 3 papers addressing mitigating bias in radiology machine learning from 3 phrases: data handling \cite{rouzrokh2022mitigating}, model development \cite{zhang2022mitigating}, and performance metrics \cite{faghani2022mitigating}. Especially in the data handling phase, they addressed 8 types of specific biases in detail, including selection bias, exclusion bias, measurement bias, recall bias, survey bias, observer bias, prejudice bias, and algorithmic bias, and also discussed debiasing methods in different phrases.

\subsubsection{Causality for Data Noise and Bias Handling} 

Doing causal effect estimation needs adjustment according to confounders.
This process is based on the strong assumption that all confounders are measurable.
However, many real-world data sets do not satisfy this assumption.
In practice, we may only observe a noisy distribution of the confounders or some proxies of the underlying confounders.
For example, income and education indices are frequently used proxies when socio-economic background is considered as a confounder.

\citet{greenlandlash2008} develop a matrix adjustment method to restore the causal effect estimation with one proxy, which is independent of the treatment and the outcome when external information of the error mechanism is given.
\citet{kuroki2014measurement} generalize the matrix adjustment method to models with multiple proxies under certain dependence settings.
However, in \cite{kuroki2014measurement}, the methods still require relatively strong assumptions, e.g., the confounder and proxies are categorical and have the same number of categories.
\citet{miao2018identifying} then further generalize the method by relaxing the above assumptions and being able to work for more general dependence settings.

To restore causal effect estimation for more general settings, \citet{louizos2017causal} develop a variational autoencoder-based method called CEVAE.
CEVAE is able to significantly relax the strong assumptions in the matrix adjustment method and deal with proxies with a different structure from the true confounder, e.g., with a continuous structure or a categorical structure but with a different number of categories.
However, \citet{rissanen2021critical} critically investigate the estimation consistency of \cite{louizos2017causal} and compare the method proposed in \cite{louizos2017causal} with the matrix adjustment method on (semi-)synthetic data.
\cite{rissanen2021critical} concludes that CEVAE may fail to output consistent estimation when the assumptions are relaxed too much.

In the context of AMR tasks, \citet{CHEN2008639} studies the influence of the time from previous hospital discharge on subsequent antimicrobial susceptibility patterns. Even though causality is not explicitly mentioned, the paper uses hospital discharge time as the instrument variable of antibiotics use because antibiotics usage decreases after hospital discharge. With the same group of patients, the method effectively reveals the causal relation between antibiotic use and antimicrobial susceptibility.

\citet{zhao2024causal} studies how to identify antibiotic resistance genes (ARGs). As ARGs have multiple properties and the annotations of these properties are unbalanced, a simple deep neural network approach may lead to the prediction of wrong ARGs due to the unseen properties associated with the sequences of ARG-encoded proteins. To mitigate such annotation biases, \cite{zhao2024causal} proposes a causality-based approach, which characterises unobserved information that generates the properties using a Gaussian Mixture Model (GMM). This approach learns the posterior of GMM, which enables the estimation of the unobserved variable. The estimation helps obtain an unbiased representation of properties in the training data. Causal-ARG further constructs a causal graph among these properties to achieve the prediction model's transferability.  

\subsubsection{Interpretability for Modele Prediction Understanding}
Interpretability methods are also used in AMR tasks to identify the properties contributing to antimicrobial resistance to detect potential risks that spurious features in the data are learned by machine learning models. 
Providing model interpretability helps answer essential questions about how and why complex machine learning systems make a decision. 
The primary goals of interpretability include the following\cite{AIExplainability}: (1) assisting human decision-making and improving trust to a certain level; (2) providing transparency to the complex optimisation pro-
cess; (3) providing information for model debugging; (4) enabling auditing and accountability.

Understanding the decision-making of a machine learning model can help mitigate both data noise and bias. \citet{weis2022direct} develop a machine learning model to directly learn from matrix-assisted laser desorption/ionization-time of flight (MALDI-TOF) mass spectra profiles of clinical isolates to predict antimicrobial resistance. The Shapley values \cite{shapley1953value} are calculated to help interpret the machine learning models. 

\citet{cavallaro2023informing} examines how interpretability methods can be used to determine patient features that have indications of the chance that the patient is susceptible to antimicrobial resistance. The work uses a gradient-boosted decision tree to predict the presence of AMR in a clinical setting based on a variety of patient data. 
Again, Shapely values are used to address the model's potential underlying dependence. The study finds that historical antibiotic prescriptions play an important role in resistance prediction. It has shown that the prescription of any antibiotic is likely to result in AMR. Explanations of prediction models either support existing hypotheses or contradict them. The latter may help identify the problem in the data. 
 
The increasing use of Shapley values as a mechanism to explain AMR prediction models faces its own challenge. As indicated by \cite{marques2024explainability}, Shapley values can provide misleading explanations of feature importance.  

Interpretability methods are also used in antibiotic discovery. \citet{wong2023discovery} use graph neural networks trained on a large dataset of compounds, predicting antibiotic activity and cytotoxicity. The work identifies the chemical substructures responsible for predicted activity using graph search under the assumption that a high prediction score must indicate there are substructures associated with the antibiotic activities, enabling the discovery of entirely new structural classes of antibiotics. This is an ad-hoc method to align the model prediction with human knowledge of substructures. 

\subsection{Mitigation Techniques in Antibiotic Discovery}

We use the antibiotic discovery task as a concrete example to illustrate data noise and bias mitigation strategies. 
We use DeepARG~\cite{arango2018deeparg} to show how noise and bias in data and model are handled.

\emph{Learning task: } This study aims to predict novel ARGs from DNA extracted directly from a wide range of environmental compartments. The input DNA sequences are first transformed to numerical representation. The representation is based on the distances of these sequences to known ARGs. The numerical representations are then fed into a deep-learning model to train the model weights. 

\emph{Data types: } The data used for the learning task are mainly from the following three databases: UNIPROT \cite{apweiler2004uniprot}, CARD \cite{jia2016card} and ARDB \cite{liu2009ardb}. UNIPROT is a comprehensive resource for protein sequence and annotation data. Its data type is a custom flat-file format that contains various fields of information about each protein entry, such as protein name, organism, gene name, sequence, functions, gene expression, references, etc. The flat file format is human-readable and can be parsed by various tools and programs. Data in CARD and ARDB are also structured with antibiotic resistance information.  

\emph{Data representativeness and data quality:} There are different ways to collect data~\cite{roh2019survey, whang2023data}. This study is built on existing datasets with improvements on data cleaning and relabelling. Data input for DeepARG includes profiled ARG from ``livestock manure, compost, wastewater treatment plants, soil, water, and other affected environments as well as human microbiome''. There is potential selection bias in the data depending on how the profiling is done. The data quality in the CARD and ARDB databases is also a problem. Some genes are assigned to wrong categories according to the authors.
However, the effort of data collection to form databases such as CARD, ARDB and UNIPROT makes data bias in terms of representativeness less significant in ARG identification compared to other AMR analytics tasks. To address the data quality problem, additional data pre-processing steps are used to consolidate antibiotics categories in the CARD and ARDB databases. Clustering and deduplication techniques are used to improve annotation quality in UNIPROT database.

\emph{Model bias:} Model biases in traditional methods are significant due to the nature of the local match methodology adopted.
Traditional methods that detect known ARG sequences from profiles are limited by the capabilities of different tools or algorithms used. The cut-offs lack a consistent definition. False negatives become a major problem in traditional methods. 
DeepARG adopts a deep learning approach to address the problem by handling a large amount of data at once so that a relatively consistent similarity measure can be derived from better data fitting. The steps of achieving this in the data layer include merging databases and handling different annotations.   

\emph{Debiasing techniques and patterns:} Even though the global effort of collecting data in UNIPROT, CARD and ARDB largely addresses the data representativeness problem, data quality remains a challenge for the ARG identification task. Major techniques used in this study include both manual and automated processes to clean the data labels by merging categories for data debiasing and use more complicated deep learning models to mitigate the model bias by avoiding setting arbitrary cut-offs with local similarity matching of ARGs. 

\subsection{A Summary of Mitigation Strategies}
Table~\ref{tab:amr_tasks} summarizes the common data problems encountered across various AMR tasks and the corresponding methods used to mitigate them. 
\textbf{Note that AMR surveillance also plays a critical role in improving data quality and representativeness by standardizing data collection and quality assurance processes \cite{world2023glass}.}
These mitigation strategies intend to enhance the reliability, fairness, and robustness of ML-centric data challenges in AMR research.
In addition to the typical mitigation approaches in data preparation and model development, this section also explores strategies for causal inference to address data noises and investigates privacy risks that may introduce biases.

\begin{table}[th!]
    \small
    \centering
    \begin{tabular}{|>{\centering\arraybackslash}p{2.1cm}|p{3.9cm}|p{8.8cm}|}
        \hline
        \textbf{AMR Tasks} & \textbf{Data Problems} & \textbf{Mitigation Methods} \\
        \hline
        \textbf{Antimicrobial Stewardship} & 
        \begin{itemize}[leftmargin=*, noitemsep, topsep=0pt]
            \item Antibiotics consumption measurements may vary among countries.
            \item Data collected by different organizations leading to inconsistencies.
        \end{itemize} & 
        \begin{enumerate}[leftmargin=*, noitemsep, topsep=0pt]
            \item Unified coding for antimicrobial classes~\cite{world2023glass}.
            \item Linking AMR and AMC in data collection to improve correlation analysis~\cite{world2023glass}.
            \item Population size weighting in analysis~\cite{ajulo2024global}.
            \item Contrastive data aggregation to enhance signals (e.g., countries with or without stewardship programs).
            \item Well-curated small data for robust analysis (e.g., international traveller data)~\cite{sridhar2021antimicrobial, dao2020infectious}.
            \item Explanation methods to infer properties lead to AMR predictions\cite{cavallaro2023informing, weis2022direct}.
        \end{enumerate} \\
        \hline
        \textbf{Antibiotics Discovery} & 
        \begin{itemize}[leftmargin=*, noitemsep, topsep=0pt]
            \item Insufficient data diversity for robust model training.
            \item Data noise and representation bias in molecular datasets.
        \end{itemize} & 
        \begin{enumerate}[leftmargin=*, noitemsep, topsep=0pt]
            \item Substantial chemical diversity guaranteed through data supplementation with natural products~\cite{stokes2020deep} or compound combination from diverse sources~\cite{das2021accelerated, li2022amplify, wong2023discovery}.
            \item Data cleaning and preprocessing including the removal of incorrect labels with cross-checking and manual screening~\cite{ma2022identification}.
            \item Dataset diversity enhanced using attribute-controlled generative models and molecular dynamics simulations~\cite{das2021accelerated}.
            \item Improved interpretability and reduced algorithmic bias with explainable AI techniques~\cite{wong2023discovery}.
            \item Causal representation learning to address the unobserved variable problem in property annotation\cite{zhao2024causal}.
        \end{enumerate} \\
        \hline
        \textbf{AMR Prediction} & 
        \begin{itemize}[leftmargin=*, noitemsep, topsep=0pt]
            \item Data noise from duplicates and incorrect annotations.
            \item Imbalanced representation of data categories.
        \end{itemize} & 
        \begin{enumerate}[leftmargin=*, noitemsep, topsep=0pt]
            \item Improved dataset quality through duplicate removal, annotation validation, and manual error correction~\cite{arango2018deeparg, gibson2015improved, li2021hmd}.
            \item Balanced datasets using resampling techniques, including oversampling and undersampling~\cite{Ren2022}.        
        \end{enumerate}\\
        \hline
        \textbf{AMR Driver Analysis} & 
        \begin{itemize}[leftmargin=*, noitemsep, topsep=0pt]
            \item Data collection bias from non-random sampling.
            \item Imbalance representation across multi-domain datasets.
        \end{itemize} & 
        \begin{enumerate}[leftmargin=*, noitemsep, topsep=0pt]
            \item Reduced data collection bias with randomized and stratified sampling methods~\cite{vikesland2019differential}.
            \item Improved data representativeness through data aggregation of large-scale authoritative datasets (e.g., WHO, ResistanceMap, World Bank) from multi-domain information (e.g., health, economic, governance)~\cite{Collignon2018, maugeri2023socio}.
            \item Enhanced minority class representation with synthetic data generation methods like SMOTE~\cite{AWASTHI2022133}.
        \end{enumerate} \\
        \hline
    \end{tabular}
    \caption{A summary of mitigation methods}
    \label{tab:amr_tasks}
\end{table}

Table \ref{table:AMR_detail} lists the details of references, including their used data, analysis methods, data issues, and data processing methods for mitigating data issues.

\subsection{Further Discussions} 


AMR research entails collecting and processing a vast volume of data, 
including sensitive patient-specific information~\cite{jensen2012mining}, which often 
requires robust privacy-preserving protocols to prevent unintended disclosures and misuse~\cite{miyaji2017privacy}.
The application of machine learning and data analytics in AMR research introduces additional risks~\cite{strobel2022data} such as membership inference, attribute inference, and data reconstruction~\cite{liu2021machine}.

However, the privacy preservation methods often lead to a reduction in the quality of data and models for analysis~\cite{hua2008survey, majeed2023quantifying}. It poses additional challenges for machine learning models developed on data processed by privacy-preserving methods~\cite{lee2017utility, qian2023efficient}, specifically, 

\begin{enumerate}
    \item Data anonymisation and perturbation: 
    While these techniques have been widely adopted to protect privacy, they can significantly degrade the data's granularity and quality, which are crucial for accurate AMR predictions~\cite{xiao2009optimal}. 
    This is because the "noise" added to protect individual identities can obscure meaningful patterns in the data, e.g., removing attribute correlation details that are vital for understanding complex AMR patterns, which reduces the predictive power of machine learning models.
    \item Data encryption and secure computation: 
    Techniques like homomorphic encryption, which allow computations on encrypted data, 
    entail tremendous computational overheads and complexities. 
    This could limit the feasibility of processing large datasets typical in AMR research~\cite{al2016secure}, 
    potentially slowing down the research progress or increasing costs prohibitively.
    \item Differentially private machine learning: 
    Integrating differential privacy into machine learning models ensures that outputs do not compromise individual privacy, 
    providing strong privacy guarantees to everyone involved. 
    However, this approach often requires a tradeoff between privacy guarantees and the accuracy of the models, 
    as higher privacy levels can lead to less precise outcomes~\cite{niinimaki2019representation}, which limits the ability of researchers to draw precise conclusions from large datasets~\cite{wang2009survey}. 
\end{enumerate}

In general, 
integrating the above privacy-preserving techniques involves a trade-off between ensuring the confidentiality of sensitive data and maintaining the quality and utility of the data for AMR research~\cite{chester2020balancing}. 
To reduce the data and model distortion, well-defined data sharing and access processes~\cite{lamb2006role} remain a practical solution. Regulatory frameworks such as the General Data Protection Regulation (GDPR) ~\cite{mitchell2022creating} are helpful in establishing trustworthiness in sensitive data handling among all parties. 
Privacy-preserving techniques, 
though critical for safeguarding patient information, 
often pollute and distort data distributions, 
thus diminishing the data's utility for AMR research.
While the pursuit of scientific breakthroughs in AMR should not come at the expense of privacy or research efficacy~\cite{jeong2022new}, extra care should be taken by AMR researchers on the data noise and bias potentially introduced by privacy preserving methods.



\section{Conclusion}

Antimicrobial resistance (AMR) poses an increasing threat to global healthcare, undermining the efficacy of modern medicine. Data-driven methodologies are frequently employed to tackle this challenge. In this paper, we investigate the implications of data noise and bias on key AMR tasks, including driver analysis, AMR prediction, Antimicrobial stewardship and new antibiotic discovery. We also review various mitigation methods. Specifically, we examine data collection, data preparation and modelling practices currently in use. We then detail approaches to reducing data noise and bias at both the data preparation and machine learning model development stages. Our work highlights the risks of producing misleading predictions in data-driven AMR tasks without proper mitigation strategies at both algorithm and procedure levels.  

Furthermore, we notice that emerging technologies, such as large language models (LLMs), provide new opportunities to address challenges in AMR research. For instance, one recent study developed an AMR-Policy GPT to help governments, researchers, and public groups (especially in low-to-middle-income countries) with AMR policy guidance for better protecting public health \cite{chen2025usingllm}. Another recent study explored to discover new drugs by using LLMs to predict the activity and toxicity of antimicrobial peptides \cite{orsi2024canllm}. 

Meanwhile, in the medical field, LLMs also prove great advancements, particularly in automating processes, improving communication, and analyzing vast datasets. For example, LLMs can automate triage, medical coding, and documentation, improving efficiency and accuracy for clinical documentation \cite{gao2023inevitable}. Also LLMs like GatorTron and GatorTronGPT can process electronic health records and clinical trial data, leading to improved diagnoses and personalized treatments \cite{gao2023inevitable,peng2023study}. Enhanced NLP capabilities in chatbots and virtual assistants facilitate better patient engagement and communication \cite{cascella2024breakthrough,gao2023inevitable}. And LLMs can create realistic synthetic data for occupational medicine, aiding in identifying new sentinal cases \cite{johnson2024what}. 

Despite these advanced models show significant potential, they also pose risks such as hallucination, misinformation and biases in training data, and privacy concerns. Ensuring accountability and ethical use is essential as these technologies evolve \cite{cascella2024breakthrough,gao2023inevitable}. 
Further researches on applying LLMs to AMR are expected to unlock their full potential and address these concerns.

\begin{landscape}
\small
\begin{longtable}[th]{|p{0.1\textwidth}|p{0.1\textwidth}|p{0.22\textwidth}|p{0.12\textwidth}| p{0.22\textwidth}| p{0.21\textwidth}|p{0.1\textwidth}|}
\caption{AMR research details of data used, analysis methods, data issues, and the mitigation methods for handling these data issues.}
\label{table:AMR_detail}\\
\hline
\textbf{Reference} & \textbf{AMR source} & \textbf{Data} & \textbf{Data issues considered} & \textbf{Mitigation} (denoising or debiasing or robust/fair modelling) & \textbf{Analytical methods} & \textbf{Tasks} \\
\hline
\citet{ALLEL2023e291}  & human-animal interaction & WHO, World Bank, and Center for Disease Dynamics Economics and Policy & Representation bias, evaluation bias & A selected set of AMR are extracted from the data sources. Keyword search of PubMed to identify associations. Leave-one-out validation. & Univariate and multivariable $\beta$ regression models. & AMR driver analysis\\
\hline
\citet{maugeri2023socio} & Human & Demographics, health, economic, and governance data (World Bank DataBank); political rights and civil liberties scores (The Freedom House); pathogens and antibiotics (The European Centre for Disease Prevention and Control (ECDC) atlas) and community consumption of antibiotics (the ESAC-Net database) & Noise & data aggregation for combining individual measures (raw data) into indices; data imputation with average values & bivariate (correlation), multivariable (regression), multivariate (clustering), and mediation analyses & AMR driver analysis\\
\hline
\citet{wong2023discovery} & Human & Original screened dataset including 39,312 compounds containing most known antibiotics, natural products, and structurally diverse molecules; Tested dataset including 12,076,365 compounds, comprising 11,277,255 from Mcule purchasable database and 799,140 compounds from a Broad Institute database. & Noise, representation bias, algorithmic bias & collect and screen compounds from diverse sources; interpretable deep learning models for better understanding and exploration.  & An explainable graph neural network (using graph-based searches to make it explainable) to train binary classifiers & new antimicrobial discovery\\
\hline
\citet{li2022amplify} & animal & A non-redundant dataset publicly available AMP sequences datasets: Antimicrobial Peptide Database (APD3) and Database of Anuran Defense Peptides (DADP) & Noise, evaluation bias & combine a non-redundant dataset from two manually curated databases; 5-fold cross-validation. & An attentive deep learning model with Bi-LSTM layer and attention layer for AMP prediction & new antimicrobial discovery\\
\hline
\citet{ma2022identification} & Human & Representative metagenomic dataset is assembled from a global human metagenomic dataset \cite{pasolli2019extensive} including more than 11,000 samples from 15 independent cohorts from year 2012 to 2019 & Noise & data preprocessing by cross-checking and screening; improving robustness by combining multiple NLP models. & Combining multiple natural language processing neural network models, including LSTM, Attention and transformer, for AMP identification & new antimicrobial discovery\\
\hline

\hline
\citet{Ren2022} & human & (1) Giessen data: whole-genome sequencing data and corresponding phenotypic information for 4 antibiotics (ciprofloxacin, cefotaxime, ceftazidime and gentamicin) for 987 E.coli strains; and (2) public data: for the same 4 antibiotics for 1509 E.coli strains. & representation bias & down-sampling & Linear regression, SVM, random forest, and CNN & AMR prediction\\
\hline
\citet{AWASTHI2022133} & human & AMR surveillance competition and the World Governance Indicators (WGI) data, the Global Burden of Disease Study (GBD) data, and the Finance data & Noise, representation bias & for missing data, variables with more than 10\% missing data are discarded, otherwise imputed by random forest; class imbalance is synthetic Minority Oversampling technique & Bayesian networks, counterfactual analysis, supervised ML (RF, SVM, LR, Naïve bayes;  & AMR driver analysis\\
\hline
\citet{weis2022direct} & Human & Create a clinical routine database: assembled MALDI-TOF mass spectra from 2016 to 2018 from more than 300,000 clinical isolates from four different diagnostic laboratories in Switzerland. The raw dataset consists of 303,195 mass spectra and 768,300 antimicrobial resistance labels and represents 803 different species of bacterial and fungal pathogens. & Noise, algorithmic bias, evaluation bias & collect and process and organize MALDI-TOF mass spectra from 4 different diagnostic laboratories in Switzerland; train AMR classifiers with 5-fold cross-validation hyperparameter search; Test stability of results with different dataset perturbations (experiments were repeated for 10 different shuffled train-test splits). & Three classifiers: logistic regression, gradient-boosted decision trees (lightGBM), and a deep neural network classifier (multilayer perceptron, MLP) & AMR prediction\\
\hline
\citet{Youn2022} &  & E. coli antibiotic resistance knowledge graph curated from 10 sources  & Noise, Representation bias & Data augmentation via knowledge inference over manually created rules; inconsistency resolution for selecting which one of two conflicting facts is more likely to be true; Do link prediction for missing links; cross-validation & iterative link prediction via 5 knowledge graph embedding methods (PRA, MLP, a stacked model that combines PRA and MLP using AdaBoost, TransE, and TransD) & AMR prediction\\
\hline
\citet{booton2021one} & Human-animal-enviroment interaction & Thailand AMR prevalence data for human, animal, environment \cite{pongpech2008prevalence,sasaki2010high,luvsansharav2011analysis,luvsansharav2012prevalence,boonyasiri2014prevalence,thamlikitkul2019contamination} & Noise, evaluation bias & Data aggregation from six separate studies (for model calibration and comparison); Latin Hypercube Sampling (a statistical method for generating random parameters from multidimensional data). & a compartmental model of ordinary differential equations (ODEs) (for describing the relationship between resistant bacteria in the three compartments humans, animals and environment) & AMR driver analysis\\
\hline

\hline
\citet{das2021accelerated} & N/A & Unlabeled molecule sequence dataset is from Uniprot database containing over 1.7M sequences; Labeled AMP sequence combines from 5 publicly available databases containing about 9000 sequences. & Noise, algorithmic bias & combine unlabeled and labelled molecule sequences from 6 publicly available databases; regularization in the encoder. & Attribute-controlled deep generative models by leveraging attribute classifiers with a rejection sampling scheme to generate molecules with the desired attributes. & new antimicrobial discovery\\
\hline
\citet{li2021hmd} & Human-environment interaction & HMD-ARG-DB, built from seven published ARG databases & Noise, evaluation bias & remove identical and duplicates in data preparation; stratified cross-validation in model development & Hierarchical multi-task deep learning (HMD-ARG): end-to-end convolutional neural network model & AMR prediction\\
\hline
\citet{stokes2020deep} & N/A & Training set: 1,760 molecules of diverse structure and function from the US Food and Drug Administration (FDA)-approved drug library, and an additional 800 natural products isolated from plant, animal, and microbial sources, are screened for growth inhibition against E. coli BW25113 and resulting in a primary training set of 2,335 molecules; Testing set: the Drug Repurposing Hub library consists of 6,111 molecules at various stages of investigation for human diseases. & Representation bias & Data supplementation with natural products to guarantee substantial chemical diversity in the training data. & a directed-message passing deep neural network model to predict the antibiotic activity of molecules (for structural consistency); Tanimoto nearest neighbour analyses (for structural diversity). & new antimicrobial discovery\\
\hline
\citet{Hur2020} & animal & VetCompass and SAVSNET & Noise & if missing data is more than 35\%, will be excluded from analysis & NLP; description analytics & AMR surveillance (for AMR stewardship)\\
\hline
\citet{nguyen2019using} & animal & Publicly available collection of 5278 Salmonella (bacterial) genome sequences & Representation bias, evaluation bias & collect over 15 years genome data; test model accuracy by 10-fold cross-validation. & Extreme gradient boosting-based regressor for prediction. & AMR prediction\\
\hline
\citet{vikesland2019differential} & Human-animal-environment interaction & publications on AMR & representation bias & data sampling may bias towards AMR concentrated values than using randomized sampling and sampling data of more confounders. & literature review & AMR driver analysis\\
\hline
\citet{gibson2015improved} & human-environment interaction & Resfams database & Noise & functionality validation and hand curated annotation & profile hidden Markov models (HMMs) & AMR prediction\\
\hline

\hline
\citet{Collignon2018} & human & AMR data (ResistanceMap, the WHO 2014 report on antimicrobial resistance, and contemporary publications), antibiotic consumption data (the IQVIA MIDAS database), and data for governance, education, gross domestic product (GDP) per capita, health-care spending, and community infrastructure (eg, sanitation) from the World Bank DataBank & Noise & data aggregation for combining individual measures (raw data) into indices & univariate analysis (correlation) and multivariable analysis (logistic regression) & AMR driver analysis\\
\hline
\citet{arango2018deeparg} & N/A & A manually-curated ARG database (DeepARG-DB) comprises 14,933 genes in total, which merged and clustered from three public databases: the Comprehensive Antibiotic Resistance Database (CARD) \cite{jia2016card}, the Antibiotic Resistance Genes Database (ARDB) \cite{liu2009ardb}, and the Universal Protein Resource (UNIPROT) database \cite{apweiler2004uniprot}. & Noise, Representation bias & Database merging with removing duplicates; AMR annotation for UNIPROT genes by sequence alignment validation; Data categorizing correction by manually inspected; A low false-positive validation for DeepARG model through PseudoARGs synthesized by randomly selected partial ARG sequences. & Two deep learning models, DeepARG-SS and DeepARG-LS, to predict ARGs from short reads and full gene length sequences, respectively. & AMR prediction\\
\hline
\citet{wang2018occurrence} & Human-environment interaction & The sequencing dataset of antibiotic-resistant bacteria (ARB) detected and organized by authors in hospital wastewater was deposited in Sequence Rad Archive. & Noise & collect hospital wastewater samples from three public hospitals; raw data preprocessed by trimming, denoising and clustering.  & Pearson's correlation was used to analyze the relationships between antibiotics and ARGs & AMR surveillance (AMR prediction)\\
\hline
\citet{xie2018antibiotics} & animal-environment & Publications and data about the manure production, use and regulations of veterinary antibiotics in China; antibiotics and antibiotic resistant genes in manure (and its composts) in China; and Effect of manure applications on the sil resistome in China & Social bias, algorithmic bias & Regulation & literature review & AMR driver analysis\\

\hline
\citet{CHEN2008639}  & Human & 789 patients enrolled from a single hospital for 1-year post-hospitalisation study & Noise & patient group assignment based on time discharged from the hospital & logistical analysis according to different post-hospitalization periods & AMR driver analysis\\
\hline
\citet{aura2023} & human & Patient antibiotic usage and AMR record data systematically collected in Australia-wise hospitals and healthcare institutes & Noise, representation bias & Removing replicated data, and adjusting for non-random sampling data. & Data distribution analysis & Surveillance (AMR driver analysis, AMR stewardship, and AMR prediction)\\
\hline
\citet{world2014antimicrobial} & human & WHO database, collecting antibiotic usage and AMR records from global hospitals & Noise, representation bias & Data cleaning & Data distribution analysis & Surveillance (AMR driver analysis, AMR stewardship, and AMR prediction)\\
\hline
\citet{wyrsch2022urban} & animal & AMR susceptibility of 425 E.coli isolates and gene-level data extracted from gull chicks’
cloacal samples at three sites in New South Wales, Australia. & Noise & Data cleaning & Data distribution analysis & Surveillance (AMR prediction)\\
\hline
\citet{blaak2015multidrug} & environment & Multi-drug resistant tests for E.coli isolates from water samples from: (1) 30 Dutch surface water bodies, and (2) 14 wastewater bodies (5 healthcare centers, 7 municipals, and 1 airport)  & Noise & Data cleaning & Kruskal Wallis test, Pearson
Chi-Square test, and Simpson’s Index of Diversity & Surveillance (AMR driver analysis)\\
\hline
\citet{zhu2013diverse} & environment & Antibiotic and metal concentrations, and gene-level AMR data are extracted from soil samples from agronomic fields applied manure-based compost near three representative large-scale swine farms in Beijing, Zhejiang, and Fujian, China, respectively. & Noise & Data cleaning  & Averaging, clustering, regression, and Canonical correspondence analysis & Surveillance (AMR driver analysis, AMR stewardship, and AMR prediction)\\
\hline
\end{longtable}
\end{landscape}

\bibliographystyle{abbrvnat}
\bibliography{main}

\begin{thebibliography}{109}
\providecommand{\natexlab}[1]{#1}
\providecommand{\url}[1]{\texttt{#1}}
\expandafter\ifx\csname urlstyle\endcsname\relax
  \providecommand{\doi}[1]{doi: #1}\else
  \providecommand{\doi}{doi: \begingroup \urlstyle{rm}\Url}\fi

\bibitem[AI(2019)]{AIExplainability}
G.~C. AI.
\newblock Ai explainability whitepaper, 2019.
\newblock URL \url{https://storage.googleapis.com/cloud-ai-whitepapers/AI%20Explainability%20Whitepaper.pdf}.
\newblock Accessed: 2024-12-14.

\bibitem[Ajulo and Awosile(2024)]{ajulo2024global}
S.~Ajulo and B.~Awosile.
\newblock Global antimicrobial resistance and use surveillance system (glass 2022): Investigating the relationship between antimicrobial resistance and antimicrobial consumption data across the participating countries.
\newblock \emph{Plos one}, 19\penalty0 (2):\penalty0 e0297921, 2024.

\bibitem[Al~Aziz et~al.(2016)Al~Aziz, Hasan, Mohammed, and Alhadidi]{al2016secure}
M.~M. Al~Aziz, M.~Z. Hasan, N.~Mohammed, and D.~Alhadidi.
\newblock Secure and efficient multiparty computation on genomic data.
\newblock In \emph{Proceedings of the 20th International Database Engineering \& Applications Symposium}, pages 278--283, 2016.

\bibitem[Allel et~al.(2023)Allel, Day, Hamilton, Lin, Furuya-Kanamori, Moore, {Van Boeckel}, Laxminarayan, and Yakob]{ALLEL2023e291}
K.~Allel, L.~Day, A.~Hamilton, L.~Lin, L.~Furuya-Kanamori, C.~E. Moore, T.~{Van Boeckel}, R.~Laxminarayan, and L.~Yakob.
\newblock Global antimicrobial-resistance drivers: an ecological country-level study at the human–animal interface.
\newblock \emph{The Lancet Planetary Health}, 7\penalty0 (4):\penalty0 e291--e303, 2023.
\newblock ISSN 2542-5196.
\newblock \doi{https://doi.org/10.1016/S2542-5196(23)00026-8}.
\newblock URL \url{https://www.sciencedirect.com/science/article/pii/S2542519623000268}.

\bibitem[Anahtar et~al.(2021)Anahtar, Yang, and Kanjilal]{Anahtar2021}
M.~N. Anahtar, J.~H. Yang, and S.~Kanjilal.
\newblock Applications of machine learning to the problem of antimicrobial resistance: an emerging model for translational research.
\newblock \emph{Journal of Clinical Microbiology}, 59, 6 2021.
\newblock ISSN 0095-1137.
\newblock \doi{10.1128/JCM.01260-20}.

\bibitem[Apweiler et~al.(2004)Apweiler, Bairoch, Wu, Barker, Boeckmann, Ferro, Gasteiger, Huang, Lopez, Magrane, et~al.]{apweiler2004uniprot}
R.~Apweiler, A.~Bairoch, C.~H. Wu, W.~C. Barker, B.~Boeckmann, S.~Ferro, E.~Gasteiger, H.~Huang, R.~Lopez, M.~Magrane, et~al.
\newblock Uniprot: the universal protein knowledgebase.
\newblock \emph{Nucleic acids research}, 32:\penalty0 D115--D119, 2004.

\bibitem[Arango-Argoty et~al.(2018)Arango-Argoty, Garner, Pruden, Heath, Vikesland, and Zhang]{arango2018deeparg}
G.~Arango-Argoty, E.~Garner, A.~Pruden, L.~S. Heath, P.~Vikesland, and L.~Zhang.
\newblock Deeparg: a deep learning approach for predicting antibiotic resistance genes from metagenomic data.
\newblock \emph{Microbiome}, 6\penalty0 (1):\penalty0 1--15, 2018.

\bibitem[Awasthi et~al.(2022)Awasthi, Rakholia, Agrawal, Dhingra, Nagori, Kaur, and Sethi]{AWASTHI2022133}
R.~Awasthi, V.~Rakholia, S.~Agrawal, L.~S. Dhingra, A.~Nagori, H.~Kaur, and T.~Sethi.
\newblock Estimating the impact of health systems factors on antimicrobial resistance in priority pathogens.
\newblock \emph{Journal of Global Antimicrobial Resistance}, 30:\penalty0 133--142, 2022.
\newblock ISSN 2213-7165.
\newblock \doi{https://doi.org/10.1016/j.jgar.2022.04.021}.
\newblock URL \url{https://www.sciencedirect.com/science/article/pii/S2213716522001011}.

\bibitem[Balasubramanian et~al.(2023)Balasubramanian, Van~Boeckel, Carmeli, Cosgrove, and Laxminarayan]{balasubramanian2023global}
R.~Balasubramanian, T.~P. Van~Boeckel, Y.~Carmeli, S.~Cosgrove, and R.~Laxminarayan.
\newblock Global incidence in hospital-associated infections resistant to antibiotics: An analysis of point prevalence surveys from 99 countries.
\newblock \emph{PLoS medicine}, 20\penalty0 (6):\penalty0 e1004178, 2023.

\bibitem[Bastarache et~al.(2022)Bastarache, Brown, Cimino, Dorr, Embi, Payne, Wilcox, and Weiner]{bastarache2022developing}
L.~Bastarache, J.~S. Brown, J.~J. Cimino, D.~A. Dorr, P.~J. Embi, P.~R. Payne, A.~B. Wilcox, and M.~G. Weiner.
\newblock Developing real-world evidence from real-world data: Transforming raw data into analytical datasets.
\newblock \emph{Learning Health Systems}, 6\penalty0 (1):\penalty0 e10293, 2022.

\bibitem[Beaudoin et~al.(2016)Beaudoin, Kabanza, Nault, and Valiquette]{beaudoin2016evaluation}
M.~Beaudoin, F.~Kabanza, V.~Nault, and L.~Valiquette.
\newblock Evaluation of a machine learning capability for a clinical decision support system to enhance antimicrobial stewardship programs.
\newblock \emph{Artificial intelligence in medicine}, 68:\penalty0 29--36, 2016.

\bibitem[Belkin et~al.(2019)Belkin, Hsu, Ma, and Mandal]{belkin2019reconciling}
M.~Belkin, D.~Hsu, S.~Ma, and S.~Mandal.
\newblock Reconciling modern machine-learning practice and the classical bias--variance trade-off.
\newblock \emph{Proceedings of the National Academy of Sciences}, 116\penalty0 (32):\penalty0 15849--15854, 2019.

\bibitem[Blaak et~al.(2015)Blaak, Lynch, Italiaander, Hamidjaja, Schets, and de~Roda~Husman]{blaak2015multidrug}
H.~Blaak, G.~Lynch, R.~Italiaander, R.~A. Hamidjaja, F.~M. Schets, and A.~M. de~Roda~Husman.
\newblock Multidrug-resistant and extended spectrum beta-lactamase-producing escherichia coli in dutch surface water and wastewater.
\newblock \emph{PLoS One}, 10\penalty0 (6):\penalty0 e0127752, 2015.

\bibitem[Boonyasiri et~al.(2014)Boonyasiri, Tangkoskul, Seenama, Saiyarin, Tiengrim, and Thamlikitkul]{boonyasiri2014prevalence}
A.~Boonyasiri, T.~Tangkoskul, C.~Seenama, J.~Saiyarin, S.~Tiengrim, and V.~Thamlikitkul.
\newblock Prevalence of antibiotic resistant bacteria in healthy adults, foods, food animals, and the environment in selected areas in thailand.
\newblock \emph{Pathogens and global health}, 108\penalty0 (5):\penalty0 235--245, 2014.

\bibitem[Booton et~al.(2021)Booton, Meeyai, Alhusein, Buller, Feil, Lambert, Mongkolsuk, Pitchforth, Reyher, Sakcamduang, et~al.]{booton2021one}
R.~D. Booton, A.~Meeyai, N.~Alhusein, H.~Buller, E.~Feil, H.~Lambert, S.~Mongkolsuk, E.~Pitchforth, K.~K. Reyher, W.~Sakcamduang, et~al.
\newblock One health drivers of antibacterial resistance: quantifying the relative impacts of human, animal and environmental use and transmission.
\newblock \emph{One Health}, 12:\penalty0 100220, 2021.

\bibitem[Cascella et~al.(2024)Cascella, Semeraro, Montomoli, Bellini, Piazza, and Bignami]{cascella2024breakthrough}
M.~Cascella, F.~Semeraro, J.~Montomoli, V.~Bellini, O.~Piazza, and E.~Bignami.
\newblock The breakthrough of large language models release for medical applications: 1-year timeline and perspectives.
\newblock \emph{Journal of Medical Systems}, 48\penalty0 (1):\penalty0 22, 2024.

\bibitem[Cavallaro et~al.(2023)Cavallaro, Moran, Collyer, McCarthy, Green, and Keeling]{cavallaro2023informing}
M.~Cavallaro, E.~Moran, B.~Collyer, N.~D. McCarthy, C.~Green, and M.~J. Keeling.
\newblock Informing antimicrobial stewardship with explainable ai.
\newblock \emph{PLOS digital health}, 2\penalty0 (1):\penalty0 e0000162, 2023.

\bibitem[Chandrasekaran et~al.(2016)Chandrasekaran, Cokol-Cakmak, Sahin, Yilancioglu, Kazan, Collins, and Cokol]{chandrasekaran2016chemogenomics}
S.~Chandrasekaran, M.~Cokol-Cakmak, N.~Sahin, K.~Yilancioglu, H.~Kazan, J.~J. Collins, and M.~Cokol.
\newblock Chemogenomics and orthology-based design of antibiotic combination therapies.
\newblock \emph{Molecular systems biology}, 12\penalty0 (5):\penalty0 872, 2016.

\bibitem[Chatterjee et~al.(2018)Chatterjee, Modarai, Naylor, Boyd, Atun, Barlow, Holmes, Johnson, and Robotham]{chatterjee2018quantifying}
A.~Chatterjee, M.~Modarai, N.~R. Naylor, S.~E. Boyd, R.~Atun, J.~Barlow, A.~H. Holmes, A.~Johnson, and J.~V. Robotham.
\newblock Quantifying drivers of antibiotic resistance in humans: a systematic review.
\newblock \emph{The Lancet Infectious Diseases}, 18\penalty0 (12):\penalty0 e368--e378, 2018.

\bibitem[Chen et~al.(2025)Chen, Li, So, Xu, Guo, Wang, Graham, and Zhu]{chen2025usingllm}
C.~Chen, S.-L. Li, A.~D. So, Y.-Y. Xu, Z.-F. Guo, X.~Wang, D.~W. Graham, and Y.-G. Zhu.
\newblock Using large language models to assist antimicrobial resistance policy development: Integrating the environment into health protection planning.
\newblock \emph{Environmental Science \& Technology}, 59\penalty0 (2):\penalty0 1243--1252, 2025.
\newblock \doi{10.1021/acs.est.4c07842}.
\newblock PMID: 39772476.

\bibitem[Chen et~al.(2008)Chen, Wu, Chang, Hsueh, Chiang, Lee, Ma, Hung, Chen, Su, Tsai, Chen, Chen, and Chen]{CHEN2008639}
S.-Y. Chen, G.~H.-M. Wu, S.-C. Chang, P.-R. Hsueh, W.-C. Chiang, C.-C. Lee, M.~H.-M. Ma, C.-C. Hung, Y.-C. Chen, C.-P. Su, K.-C. Tsai, T.~H.-H. Chen, S.-C. Chen, and W.-J. Chen.
\newblock Bacteremia in previously hospitalized patients: Prolonged effect from previous hospitalization and risk factors for antimicrobial-resistant bacterial infections.
\newblock \emph{Annals of Emergency Medicine}, 51\penalty0 (5):\penalty0 639--646, 2008.
\newblock ISSN 0196-0644.
\newblock \doi{https://doi.org/10.1016/j.annemergmed.2007.12.022}.
\newblock URL \url{https://www.sciencedirect.com/science/article/pii/S0196064407019002}.

\bibitem[Chester et~al.(2020)Chester, Koh, Wicker, Sun, and Lee]{chester2020balancing}
A.~Chester, Y.~S. Koh, J.~Wicker, Q.~Sun, and J.~Lee.
\newblock Balancing utility and fairness against privacy in medical data.
\newblock In \emph{2020 IEEE Symposium Series on Computational Intelligence (SSCI)}, pages 1226--1233. IEEE, 2020.

\bibitem[Chiu and Ong(2019)]{Chiu2019}
J.~K.~H. Chiu and R.~T.-H. Ong.
\newblock Argdit: a validation and integration toolkit for antimicrobial resistance gene databases.
\newblock \emph{Bioinformatics}, 35:\penalty0 2466--2474, 7 2019.
\newblock ISSN 1367-4803.
\newblock \doi{10.1093/bioinformatics/bty987}.

\bibitem[Collignon et~al.(2018)Collignon, Beggs, Walsh, Gandra, and Laxminarayan]{Collignon2018}
P.~Collignon, J.~J. Beggs, T.~R. Walsh, S.~Gandra, and R.~Laxminarayan.
\newblock Anthropological and socioeconomic factors contributing to global antimicrobial resistance: a univariate and multivariable analysis.
\newblock \emph{The Lancet Planetary Health}, 2:\penalty0 e398--e405, 9 2018.
\newblock ISSN 25425196.
\newblock \doi{10.1016/S2542-5196(18)30186-4}.

\bibitem[Crofts et~al.(2017)Crofts, Gasparrini, and Dantas]{crofts2017next}
T.~S. Crofts, A.~J. Gasparrini, and G.~Dantas.
\newblock Next-generation approaches to understand and combat the antibiotic resistome.
\newblock \emph{Nature Reviews Microbiology}, 15\penalty0 (7):\penalty0 422--434, 2017.

\bibitem[Dao et~al.(2020)Dao, Ly, Magmoun, Canard, Drali, Fenollar, Ninove, Raoult, Parola, Courjon, et~al.]{dao2020infectious}
T.~L. Dao, T.~D.~A. Ly, A.~Magmoun, N.~Canard, T.~Drali, F.~Fenollar, L.~Ninove, D.~Raoult, P.~Parola, J.~Courjon, et~al.
\newblock Infectious disease symptoms and microbial carriage among french medical students travelling abroad: a prospective study.
\newblock \emph{Travel Medicine and Infectious Disease}, 34:\penalty0 101548, 2020.

\bibitem[Das et~al.(2021)Das, Sercu, Wadhawan, Padhi, Gehrmann, Cipcigan, Chenthamarakshan, Strobelt, Dos~Santos, Chen, et~al.]{das2021accelerated}
P.~Das, T.~Sercu, K.~Wadhawan, I.~Padhi, S.~Gehrmann, F.~Cipcigan, V.~Chenthamarakshan, H.~Strobelt, C.~Dos~Santos, P.-Y. Chen, et~al.
\newblock Accelerated antimicrobial discovery via deep generative models and molecular dynamics simulations.
\newblock \emph{Nature Biomedical Engineering}, 5\penalty0 (6):\penalty0 613--623, 2021.

\bibitem[Dyar et~al.(2017)Dyar, Huttner, Schouten, and Pulcini]{DYAR2017793}
O.~Dyar, B.~Huttner, J.~Schouten, and C.~Pulcini.
\newblock What is antimicrobial stewardship?
\newblock \emph{Clinical Microbiology and Infection}, 23\penalty0 (11):\penalty0 793--798, 2017.
\newblock ISSN 1198-743X.
\newblock \doi{https://doi.org/10.1016/j.cmi.2017.08.026}.
\newblock URL \url{https://www.sciencedirect.com/science/article/pii/S1198743X17304895}.

\bibitem[Faghani et~al.(2022)Faghani, Khosravi, Zhang, Moassefi, Jagtap, Nugen, Vahdati, Kuanar, Rassoulinejad-Mousavi, Singh, et~al.]{faghani2022mitigating}
S.~Faghani, B.~Khosravi, K.~Zhang, M.~Moassefi, J.~M. Jagtap, F.~Nugen, S.~Vahdati, S.~P. Kuanar, S.~M. Rassoulinejad-Mousavi, Y.~Singh, et~al.
\newblock Mitigating bias in radiology machine learning: 3. performance metrics.
\newblock \emph{Radiology: Artificial Intelligence}, 4\penalty0 (5):\penalty0 e220061, 2022.

\bibitem[Frost et~al.(2021)Frost, Kapoor, Craig, Liu, and Laxminarayan]{FROST2021222}
I.~Frost, G.~Kapoor, J.~Craig, D.~Liu, and R.~Laxminarayan.
\newblock Status, challenges and gaps in antimicrobial resistance surveillance around the world.
\newblock \emph{Journal of Global Antimicrobial Resistance}, 25:\penalty0 222--226, 2021.
\newblock ISSN 2213-7165.
\newblock \doi{https://doi.org/10.1016/j.jgar.2021.03.016}.
\newblock URL \url{https://www.sciencedirect.com/science/article/pii/S2213716521000825}.

\bibitem[Gao et~al.(2023)Gao, Baptista-Hon, and Zhang]{gao2023inevitable}
Y.~Gao, D.~T. Baptista-Hon, and K.~Zhang.
\newblock The inevitable transformation of medicine and research by large language models: the possibilities and pitfalls.
\newblock \emph{MEDCOMM-Future Medicine}, 2\penalty0 (2), 2023.

\bibitem[Gianfrancesco et~al.(2018)Gianfrancesco, Tamang, Yazdany, and Schmajuk]{gianfrancesco2018potential}
M.~A. Gianfrancesco, S.~Tamang, J.~Yazdany, and G.~Schmajuk.
\newblock Potential biases in machine learning algorithms using electronic health record data.
\newblock \emph{JAMA internal medicine}, 178\penalty0 (11):\penalty0 1544--1547, 2018.

\bibitem[Gibson et~al.(2015)Gibson, Forsberg, and Dantas]{gibson2015improved}
M.~K. Gibson, K.~J. Forsberg, and G.~Dantas.
\newblock Improved annotation of antibiotic resistance determinants reveals microbial resistomes cluster by ecology.
\newblock \emph{The ISME journal}, 9\penalty0 (1):\penalty0 207--216, 2015.

\bibitem[Government(2023)]{aura2023}
A.~Government.
\newblock \emph{Fifth Australian report on antimicrobial use and resistance in human health (AURA 2023)}.
\newblock Australian Commission On Safety And Quality In Health Care, 2023.

\bibitem[Greenland and Lash(2008)]{greenlandlash2008}
S.~Greenland and T.~Lash.
\newblock Bias analysis.
\newblock In S.~G. K.~Rothman and T.~Lash, editors, \emph{Modern Epidemiology}, pages 345--380. Philadelphia: Lippincott Williams and Wilkins, 3 edition, 2008.

\bibitem[Hua and Pei(2008)]{hua2008survey}
M.~Hua and J.~Pei.
\newblock A survey of utility-based privacy-preserving data transformation methods.
\newblock \emph{Privacy-Preserving Data Mining: Models and Algorithms}, pages 207--237, 2008.

\bibitem[Hur et~al.(2020)Hur, Hardefeldt, Verspoor, Baldwin, and Gilkerson]{Hur2020}
B.~A. Hur, L.~Y. Hardefeldt, K.~M. Verspoor, T.~Baldwin, and J.~R. Gilkerson.
\newblock Describing the antimicrobial usage patterns of companion animal veterinary practices; free text analysis of more than 4.4 million consultation records.
\newblock \emph{PLOS ONE}, 15:\penalty0 e0230049, 3 2020.
\newblock ISSN 1932-6203.
\newblock \doi{10.1371/journal.pone.0230049}.
\newblock URL \url{https://dx.plos.org/10.1371/journal.pone.0230049}.

\bibitem[Iskandar et~al.(2021)Iskandar, Molinier, Hallit, Sartelli, Hardcastle, Haque, Lugova, Dhingra, Sharma, Islam, et~al.]{iskandar2021surveillance}
K.~Iskandar, L.~Molinier, S.~Hallit, M.~Sartelli, T.~C. Hardcastle, M.~Haque, H.~Lugova, S.~Dhingra, P.~Sharma, S.~Islam, et~al.
\newblock Surveillance of antimicrobial resistance in low-and middle-income countries: a scattered picture.
\newblock \emph{Antimicrobial Resistance \& Infection Control}, 10:\penalty0 1--19, 2021.

\bibitem[Jensen et~al.(2012)Jensen, Jensen, and Brunak]{jensen2012mining}
P.~B. Jensen, L.~J. Jensen, and S.~Brunak.
\newblock Mining electronic health records: towards better research applications and clinical care.
\newblock \emph{Nature Reviews Genetics}, 13\penalty0 (6):\penalty0 395--405, 2012.

\bibitem[Jeong et~al.(2022)Jeong, Kim, and Im]{jeong2022new}
D.~Jeong, J.~H. Kim, and J.~Im.
\newblock A new global measure to simultaneously evaluate data utility and privacy risk.
\newblock \emph{IEEE Transactions on Information Forensics and Security}, 18:\penalty0 715--729, 2022.

\bibitem[Jia et~al.(2016)Jia, Raphenya, Alcock, Waglechner, Guo, Tsang, Lago, Dave, Pereira, Sharma, et~al.]{jia2016card}
B.~Jia, A.~R. Raphenya, B.~Alcock, N.~Waglechner, P.~Guo, K.~K. Tsang, B.~A. Lago, B.~M. Dave, S.~Pereira, A.~N. Sharma, et~al.
\newblock Card 2017: expansion and model-centric curation of the comprehensive antibiotic resistance database.
\newblock \emph{Nucleic acids research}, page gkw1004, 2016.

\bibitem[Johnson(2024)]{johnson2024what}
M.~W. Johnson.
\newblock P-320 what does a large language model know about the prevalence of occupationally-related medical conditions? experiments with synthetic and real occupational medicine data.
\newblock \emph{Occupational Medicine}, 74\penalty0 (Supplement\_1), 07 2024.
\newblock ISSN 0962-7480.
\newblock \doi{10.1093/occmed/kqae023.0927}.
\newblock URL \url{https://doi.org/10.1093/occmed/kqae023.0927}.

\bibitem[Kanjilal et~al.(2020)Kanjilal, Oberst, Boominathan, Zhou, Hooper, and Sontag]{Kanjilal2020}
S.~Kanjilal, M.~Oberst, S.~Boominathan, H.~Zhou, D.~C. Hooper, and D.~Sontag.
\newblock A decision algorithm to promote outpatient antimicrobial stewardship for uncomplicated urinary tract infection.
\newblock \emph{Science Translational Medicine}, 12\penalty0 (568):\penalty0 eaay5067, 2020.
\newblock \doi{10.1126/scitranslmed.aay5067}.

\bibitem[Kuroki and Pearl(2014)]{kuroki2014measurement}
M.~Kuroki and J.~Pearl.
\newblock Measurement bias and effect restoration in causal inference.
\newblock \emph{Biometrika}, 101\penalty0 (2):\penalty0 423--437, 2014.

\bibitem[Lamb et~al.(2006)Lamb, Power, Walker, and Compton]{lamb2006role}
P.~Lamb, R.~Power, G.~Walker, and M.~Compton.
\newblock Role-based access control for data service integration.
\newblock In \emph{Proceedings of the 3rd ACM workshop on Secure web services}, pages 3--12, 2006.

\bibitem[Lee et~al.(2017)Lee, Kim, Kim, and Chung]{lee2017utility}
H.~Lee, S.~Kim, J.~W. Kim, and Y.~D. Chung.
\newblock Utility-preserving anonymization for health data publishing.
\newblock \emph{BMC medical informatics and decision making}, 17:\penalty0 1--12, 2017.

\bibitem[Lee et~al.(2022)Lee, Shin, Kim, Lee, Yoon, Lee, Kim, Kim, and Lee]{lee2022hybrid}
S.~Lee, J.~Shin, H.~S. Kim, M.~J. Lee, J.~M. Yoon, S.~Lee, Y.~Kim, J.-Y. Kim, and S.~Lee.
\newblock Hybrid method incorporating a rule-based approach and deep learning for prescription error prediction.
\newblock \emph{Drug safety}, 45\penalty0 (1):\penalty0 27--35, 2022.

\bibitem[Leibovici et~al.(2013)Leibovici, Kariv, and Paul]{Leibovici2013}
L.~Leibovici, G.~Kariv, and M.~Paul.
\newblock {Long-term survival in patients included in a randomized controlled trial of TREAT, a decision support system for antibiotic treatment}.
\newblock \emph{Journal of Antimicrobial Chemotherapy}, 68\penalty0 (11):\penalty0 2664--2666, 06 2013.
\newblock ISSN 0305-7453.
\newblock \doi{10.1093/jac/dkt222}.
\newblock URL \url{https://doi.org/10.1093/jac/dkt222}.

\bibitem[Li et~al.(2022)Li, Sutherland, Hammond, Yang, Taho, Bergman, Houston, Warren, Wong, Hoang, et~al.]{li2022amplify}
C.~Li, D.~Sutherland, S.~A. Hammond, C.~Yang, F.~Taho, L.~Bergman, S.~Houston, R.~L. Warren, T.~Wong, L.~Hoang, et~al.
\newblock Amplify: attentive deep learning model for discovery of novel antimicrobial peptides effective against who priority pathogens.
\newblock \emph{BMC genomics}, 23\penalty0 (1):\penalty0 1--15, 2022.

\bibitem[Li et~al.(2021)Li, Xu, Han, Cao, Umarov, Yan, Fan, Chen, Duarte, Li, et~al.]{li2021hmd}
Y.~Li, Z.~Xu, W.~Han, H.~Cao, R.~Umarov, A.~Yan, M.~Fan, H.~Chen, C.~M. Duarte, L.~Li, et~al.
\newblock Hmd-arg: hierarchical multi-task deep learning for annotating antibiotic resistance genes.
\newblock \emph{Microbiome}, 9:\penalty0 1--12, 2021.

\bibitem[Liguori et~al.(2022)Liguori, Keenum, Davis, Calarco, Milligan, Harwood, and Pruden]{liguori2022antimicrobial}
K.~Liguori, I.~Keenum, B.~C. Davis, J.~Calarco, E.~Milligan, V.~J. Harwood, and A.~Pruden.
\newblock Antimicrobial resistance monitoring of water environments: a framework for standardized methods and quality control.
\newblock \emph{Environmental science \& technology}, 56\penalty0 (13):\penalty0 9149--9160, 2022.

\bibitem[Liu and Pop(2009)]{liu2009ardb}
B.~Liu and M.~Pop.
\newblock Ardb---antibiotic resistance genes database.
\newblock \emph{Nucleic acids research}, 37:\penalty0 D443--D447, 2009.

\bibitem[Liu et~al.(2021)Liu, Ding, Shaham, Rahayu, Farokhi, and Lin]{liu2021machine}
B.~Liu, M.~Ding, S.~Shaham, W.~Rahayu, F.~Farokhi, and Z.~Lin.
\newblock When machine learning meets privacy: A survey and outlook.
\newblock \emph{ACM Computing Surveys (CSUR)}, 54\penalty0 (2):\penalty0 1--36, 2021.

\bibitem[Louizos et~al.(2017)Louizos, Shalit, Mooij, Sontag, Zemel, and Welling]{louizos2017causal}
C.~Louizos, U.~Shalit, J.~M. Mooij, D.~Sontag, R.~Zemel, and M.~Welling.
\newblock Causal effect inference with deep latent-variable models.
\newblock \emph{Advances in neural information processing systems}, 30, 2017.

\bibitem[Luvsansharav et~al.(2011)Luvsansharav, Hirai, Niki, Sasaki, Makimoto, Komalamisra, Maipanich, Kusolsuk, Sa-Nguankiat, Pubampen, et~al.]{luvsansharav2011analysis}
U.-O. Luvsansharav, I.~Hirai, M.~Niki, T.~Sasaki, K.~Makimoto, C.~Komalamisra, W.~Maipanich, T.~Kusolsuk, S.~Sa-Nguankiat, S.~Pubampen, et~al.
\newblock Analysis of risk factors for a high prevalence of extended-spectrum $\beta$-lactamase-producing enterobacteriaceae in asymptomatic individuals in rural thailand.
\newblock \emph{Journal of medical microbiology}, 60\penalty0 (5):\penalty0 619--624, 2011.

\bibitem[Luvsansharav et~al.(2012)Luvsansharav, Hirai, Nakata, Imura, Yamauchi, Niki, Komalamisra, Kusolsuk, and Yamamoto]{luvsansharav2012prevalence}
U.-O. Luvsansharav, I.~Hirai, A.~Nakata, K.~Imura, K.~Yamauchi, M.~Niki, C.~Komalamisra, T.~Kusolsuk, and Y.~Yamamoto.
\newblock Prevalence of and risk factors associated with faecal carriage of ctx-m $\beta$-lactamase-producing enterobacteriaceae in rural thai communities.
\newblock \emph{Journal of antimicrobial chemotherapy}, 67\penalty0 (7):\penalty0 1769--1774, 2012.

\bibitem[Ma et~al.(2022)Ma, Guo, Xia, Zhang, Liu, Yu, Tang, Tong, Wang, Ye, et~al.]{ma2022identification}
Y.~Ma, Z.~Guo, B.~Xia, Y.~Zhang, X.~Liu, Y.~Yu, N.~Tang, X.~Tong, M.~Wang, X.~Ye, et~al.
\newblock Identification of antimicrobial peptides from the human gut microbiome using deep learning.
\newblock \emph{Nature Biotechnology}, 40\penalty0 (6):\penalty0 921--931, 2022.

\bibitem[Majeed and Hwang(2023)]{majeed2023quantifying}
A.~Majeed and S.~O. Hwang.
\newblock Quantifying the vulnerability of attributes for effective privacy preservation using machine learning.
\newblock \emph{IEEE Access}, 11:\penalty0 4400--4411, 2023.

\bibitem[Marques-Silva and Huang(2024)]{marques2024explainability}
J.~Marques-Silva and X.~Huang.
\newblock Explainability is not a game.
\newblock \emph{Communications of the ACM}, 67\penalty0 (7):\penalty0 66--75, 2024.

\bibitem[Maugeri et~al.(2023)Maugeri, Barchitta, Puglisi, and Agodi]{maugeri2023socio}
A.~Maugeri, M.~Barchitta, F.~Puglisi, and A.~Agodi.
\newblock Socio-economic, governance and health indicators shaping antimicrobial resistance: an ecological analysis of 30 european countries.
\newblock \emph{Globalization and Health}, 19\penalty0 (1):\penalty0 1--12, 2023.

\bibitem[Miao et~al.(2018)Miao, Geng, and Tchetgen~Tchetgen]{miao2018identifying}
W.~Miao, Z.~Geng, and E.~J. Tchetgen~Tchetgen.
\newblock Identifying causal effects with proxy variables of an unmeasured confounder.
\newblock \emph{Biometrika}, 105\penalty0 (4):\penalty0 987--993, 2018.

\bibitem[Mitchell et~al.(2022)Mitchell, O’Neill, and King]{mitchell2022creating}
J.~Mitchell, A.~J. O’Neill, and R.~King.
\newblock Creating a framework to align antimicrobial resistance (amr) research with the global guidance: a viewpoint.
\newblock \emph{Journal of Antimicrobial Chemotherapy}, 77\penalty0 (9):\penalty0 2315--2320, 2022.

\bibitem[Miyaji et~al.(2017)Miyaji, Nakasho, and Nishida]{miyaji2017privacy}
A.~Miyaji, K.~Nakasho, and S.~Nishida.
\newblock Privacy-preserving integration of medical data: a practical multiparty private set intersection.
\newblock \emph{Journal of medical systems}, 41:\penalty0 1--10, 2017.

\bibitem[Nguyen et~al.(2019)Nguyen, Long, McDermott, Olsen, Olson, Stevens, Tyson, Zhao, and Davis]{nguyen2019using}
M.~Nguyen, S.~W. Long, P.~F. McDermott, R.~J. Olsen, R.~Olson, R.~L. Stevens, G.~H. Tyson, S.~Zhao, and J.~J. Davis.
\newblock Using machine learning to predict antimicrobial mics and associated genomic features for nontyphoidal salmonella.
\newblock \emph{Journal of Clinical Microbiology}, 57\penalty0 (2):\penalty0 e01260--18, 2019.

\bibitem[Niinim{\"a}ki et~al.(2019)Niinim{\"a}ki, Heikkil{\"a}, Honkela, and Kaski]{niinimaki2019representation}
T.~Niinim{\"a}ki, M.~A. Heikkil{\"a}, A.~Honkela, and S.~Kaski.
\newblock Representation transfer for differentially private drug sensitivity prediction.
\newblock \emph{Bioinformatics}, 35\penalty0 (14):\penalty0 i218--i224, 2019.

\bibitem[Obermeyer et~al.(2019)Obermeyer, Powers, Vogeli, and Mullainathan]{obermeyer2019dissecting}
Z.~Obermeyer, B.~Powers, C.~Vogeli, and S.~Mullainathan.
\newblock Dissecting racial bias in an algorithm used to manage the health of populations.
\newblock \emph{Science}, 366\penalty0 (6464):\penalty0 447--453, 2019.

\bibitem[Oonsivilai et~al.(2018)Oonsivilai, Mo, Luangasanatip, Lubell, Miliya, Tan, Loeuk, Turner, and Cooper]{oonsivilai2018using}
M.~Oonsivilai, Y.~Mo, N.~Luangasanatip, Y.~Lubell, T.~Miliya, P.~Tan, L.~Loeuk, P.~Turner, and B.~S. Cooper.
\newblock Using machine learning to guide targeted and locally-tailored empiric antibiotic prescribing in a children's hospital in cambodia.
\newblock \emph{Wellcome open research}, 3, 2018.

\bibitem[Organization(2015)]{10665-193736}
W.~H. Organization.
\newblock \emph{Global action plan on antimicrobial resistance}.
\newblock World Health Organization, 2015.

\bibitem[Organization(2021)]{world2021onehealth}
W.~H. Organization.
\newblock \emph{WHO integrated global surveillance on ESBL-producing E. coli using a “One Health” approach: Implementation and opportunities}.
\newblock World Health Organization, 2021.

\bibitem[Organization(2023)]{world2023glass}
W.~H. Organization.
\newblock \emph{GLASS manual for antimicrobial resistance surveillance in common bacteria causing human infection}.
\newblock World Health Organization, 2023.

\bibitem[Organization et~al.(2014)]{world2014antimicrobial}
W.~H. Organization et~al.
\newblock \emph{Antimicrobial resistance: global report on surveillance}.
\newblock World Health Organization, 2014.

\bibitem[Orsi and Reymond(2024)]{orsi2024canllm}
M.~Orsi and J.-L. Reymond.
\newblock Can large language models predict antimicrobial peptide activity and toxicity?
\newblock \emph{RSC Medicinal Chemistry}, 15:\penalty0 2030--2036, 2024.
\newblock \doi{10.1039/D4MD00159A}.

\bibitem[Pasolli et~al.(2019)Pasolli, Asnicar, Manara, Zolfo, Karcher, Armanini, Beghini, Manghi, Tett, Ghensi, et~al.]{pasolli2019extensive}
E.~Pasolli, F.~Asnicar, S.~Manara, M.~Zolfo, N.~Karcher, F.~Armanini, F.~Beghini, P.~Manghi, A.~Tett, P.~Ghensi, et~al.
\newblock Extensive unexplored human microbiome diversity revealed by over 150,000 genomes from metagenomes spanning age, geography, and lifestyle.
\newblock \emph{Cell}, 176\penalty0 (3):\penalty0 649--662, 2019.

\bibitem[Paxton~C(2013)]{Paxton2013}
S.~S. Paxton~C, Niculescu-Mizil~A.
\newblock Developing predictive models using electronic medical records: challenges and pitfalls.
\newblock \emph{AMIA Annu Symp Proc.}, pages 1109--1115, 2013.

\bibitem[Peng et~al.(2023)Peng, Yang, Chen, Smith, PourNejatian, Costa, Martin, Flores, Zhang, Magoc, et~al.]{peng2023study}
C.~Peng, X.~Yang, A.~Chen, K.~E. Smith, N.~PourNejatian, A.~B. Costa, C.~Martin, M.~G. Flores, Y.~Zhang, T.~Magoc, et~al.
\newblock A study of generative large language model for medical research and healthcare.
\newblock \emph{NPJ digital medicine}, 6\penalty0 (1):\penalty0 210, 2023.

\bibitem[Pfohl et~al.(2019)Pfohl, Marafino, Coulet, Rodriguez, Palaniappan, and Shah]{pfohl2019creating}
S.~Pfohl, B.~Marafino, A.~Coulet, F.~Rodriguez, L.~Palaniappan, and N.~H. Shah.
\newblock Creating fair models of atherosclerotic cardiovascular disease risk.
\newblock In \emph{Proceedings of the 2019 AAAI/ACM Conference on AI, Ethics, and Society}, pages 271--278, 2019.

\bibitem[Pongpech et~al.(2008)Pongpech, Naenna, Taipobsakul, Tribuddharat, and Srifuengfung]{pongpech2008prevalence}
P.~Pongpech, P.~Naenna, Y.~Taipobsakul, C.~Tribuddharat, and S.~Srifuengfung.
\newblock Prevalence of extended-spectrum beta-lactamase and class 1 integron integrase gene inti1 in escherichia coli from thai patients and healthy adults.
\newblock \emph{Southeast Asian journal of tropical medicine and public health}, 39\penalty0 (3):\penalty0 425, 2008.

\bibitem[Poulain et~al.(2023)Poulain, Bin~Tarek, and Beheshti]{poulain2023improving}
R.~Poulain, M.~F. Bin~Tarek, and R.~Beheshti.
\newblock Improving fairness in ai models on electronic health records: The case for federated learning methods.
\newblock In \emph{Proceedings of the 2023 ACM Conference on Fairness, Accountability, and Transparency}, pages 1599--1608, 2023.

\bibitem[Qian et~al.(2023)Qian, Li, and Zhou]{qian2023efficient}
X.~Qian, X.~Li, and Z.~Zhou.
\newblock An efficient privacy-preserving approach for data publishing.
\newblock \emph{Journal of Ambient Intelligence and Humanized Computing}, 14\penalty0 (3):\penalty0 2077--2093, 2023.

\bibitem[Ren et~al.(2022)Ren, Chakraborty, Doijad, Falgenhauer, Falgenhauer, Goesmann, Hauschild, Schwengers, and Heider]{Ren2022}
Y.~Ren, T.~Chakraborty, S.~Doijad, L.~Falgenhauer, J.~Falgenhauer, A.~Goesmann, A.-C. Hauschild, O.~Schwengers, and D.~Heider.
\newblock Prediction of antimicrobial resistance based on whole-genome sequencing and machine learning.
\newblock \emph{Bioinformatics}, 38:\penalty0 325--334, 1 2022.
\newblock ISSN 1367-4803.
\newblock \doi{10.1093/bioinformatics/btab681}.

\bibitem[Rissanen and Marttinen(2021)]{rissanen2021critical}
S.~Rissanen and P.~Marttinen.
\newblock A critical look at the consistency of causal estimation with deep latent variable models.
\newblock \emph{Advances in Neural Information Processing Systems}, 34:\penalty0 4207--4217, 2021.

\bibitem[Roh et~al.(2019)Roh, Heo, and Whang]{roh2019survey}
Y.~Roh, G.~Heo, and S.~E. Whang.
\newblock A survey on data collection for machine learning: a big data-ai integration perspective.
\newblock \emph{IEEE Transactions on Knowledge and Data Engineering}, 33\penalty0 (4):\penalty0 1328--1347, 2019.

\bibitem[Rouzrokh et~al.(2022)Rouzrokh, Khosravi, Faghani, Moassefi, Vera~Garcia, Singh, Zhang, Conte, and Erickson]{rouzrokh2022mitigating}
P.~Rouzrokh, B.~Khosravi, S.~Faghani, M.~Moassefi, D.~V. Vera~Garcia, Y.~Singh, K.~Zhang, G.~M. Conte, and B.~J. Erickson.
\newblock Mitigating bias in radiology machine learning: 1. data handling.
\newblock \emph{Radiology: Artificial Intelligence}, 4\penalty0 (5):\penalty0 e210290, 2022.

\bibitem[Sasaki et~al.(2010)Sasaki, Hirai, Niki, Nakamura, Komalamisra, Maipanich, Kusolsuk, Sa-Nguankiat, Pubampen, and Yamamoto]{sasaki2010high}
T.~Sasaki, I.~Hirai, M.~Niki, T.~Nakamura, C.~Komalamisra, W.~Maipanich, T.~Kusolsuk, S.~Sa-Nguankiat, S.~Pubampen, and Y.~Yamamoto.
\newblock High prevalence of ctx-m $\beta$-lactamase-producing enterobacteriaceae in stool specimens obtained from healthy individuals in thailand.
\newblock \emph{Journal of Antimicrobial Chemotherapy}, 65\penalty0 (4):\penalty0 666--668, 2010.

\bibitem[Shapley(1953)]{shapley1953value}
L.~S. Shapley.
\newblock A value for n-person games.
\newblock \emph{Contribution to the Theory of Games}, 2, 1953.

\bibitem[Sridhar et~al.(2021)Sridhar, Turbett, Harris, and LaRocque]{sridhar2021antimicrobial}
S.~Sridhar, S.~E. Turbett, J.~B. Harris, and R.~C. LaRocque.
\newblock Antimicrobial-resistant bacteria in international travelers.
\newblock \emph{Current Opinion in Infectious Diseases}, 34\penalty0 (5):\penalty0 423--431, 2021.

\bibitem[Stokes et~al.(2020)Stokes, Yang, Swanson, Jin, Cubillos-Ruiz, Donghia, MacNair, French, Carfrae, Bloom-Ackermann, et~al.]{stokes2020deep}
J.~M. Stokes, K.~Yang, K.~Swanson, W.~Jin, A.~Cubillos-Ruiz, N.~M. Donghia, C.~R. MacNair, S.~French, L.~A. Carfrae, Z.~Bloom-Ackermann, et~al.
\newblock A deep learning approach to antibiotic discovery.
\newblock \emph{Cell}, 180\penalty0 (4):\penalty0 688--702, 2020.

\bibitem[Strobel and Shokri(2022)]{strobel2022data}
M.~Strobel and R.~Shokri.
\newblock Data privacy and trustworthy machine learning.
\newblock \emph{IEEE Security \& Privacy}, 20\penalty0 (5):\penalty0 44--49, 2022.

\bibitem[Thamlikitkul et~al.(2019)Thamlikitkul, Tiengrim, Thamthaweechok, Buranapakdee, and Chiemchaisri]{thamlikitkul2019contamination}
V.~Thamlikitkul, S.~Tiengrim, N.~Thamthaweechok, P.~Buranapakdee, and W.~Chiemchaisri.
\newblock Contamination by antibiotic-resistant bacteria in selected environments in thailand.
\newblock \emph{International Journal of Environmental Research and Public Health}, 16\penalty0 (19):\penalty0 3753, 2019.

\bibitem[{UK Department of Health}(2014)]{ukamrmap2014}
{UK Department of Health}.
\newblock Department of health antimicrobial resistance (amr) systems map, Dec 2014.
\newblock URL \url{https://www.gov.uk/government/publications/antimicrobial-resistance-amr-systems-map}.

\bibitem[Van~Giffen et~al.(2022)Van~Giffen, Herhausen, and Fahse]{van2022overcoming}
B.~Van~Giffen, D.~Herhausen, and T.~Fahse.
\newblock Overcoming the pitfalls and perils of algorithms: A classification of machine learning biases and mitigation methods.
\newblock \emph{Journal of Business Research}, 144:\penalty0 93--106, 2022.

\bibitem[Vikesland et~al.(2019)Vikesland, Garner, Gupta, Kang, Maile-Moskowitz, and Zhu]{vikesland2019differential}
P.~Vikesland, E.~Garner, S.~Gupta, S.~Kang, A.~Maile-Moskowitz, and N.~Zhu.
\newblock Differential drivers of antimicrobial resistance across the world.
\newblock \emph{Accounts of chemical research}, 52\penalty0 (4):\penalty0 916--924, 2019.

\bibitem[Walsh and Wencewicz(2020)]{walsh2020antibiotics}
C.~Walsh and T.~Wencewicz.
\newblock \emph{Antibiotics: Challenges, Mechanisms, Opportunities}.
\newblock ASM Books. Wiley, 2020.
\newblock ISBN 9781555819316.
\newblock URL \url{https://books.google.com.au/books?id=oEL2DwAAQBAJ}.

\bibitem[Wang et~al.(2009)Wang, Luo, Zhao, and Le]{wang2009survey}
J.~Wang, Y.~Luo, Y.~Zhao, and J.~Le.
\newblock A survey on privacy preserving data mining.
\newblock In \emph{2009 First International Workshop on Database Technology and Applications}, pages 111--114. IEEE, 2009.

\bibitem[Wang et~al.(2018)Wang, Wang, and Yang]{wang2018occurrence}
Q.~Wang, P.~Wang, and Q.~Yang.
\newblock Occurrence and diversity of antibiotic resistance in untreated hospital wastewater.
\newblock \emph{Science of the Total Environment}, 621:\penalty0 990--999, 2018.

\bibitem[Weis et~al.(2022)Weis, Cu{\'e}nod, Rieck, Dubuis, Graf, Lang, Oberle, Brackmann, S{\o}gaard, Osthoff, et~al.]{weis2022direct}
C.~Weis, A.~Cu{\'e}nod, B.~Rieck, O.~Dubuis, S.~Graf, C.~Lang, M.~Oberle, M.~Brackmann, K.~K. S{\o}gaard, M.~Osthoff, et~al.
\newblock Direct antimicrobial resistance prediction from clinical maldi-tof mass spectra using machine learning.
\newblock \emph{Nature medicine}, pages 1--11, 2022.

\bibitem[Whang et~al.(2023)Whang, Roh, Song, and Lee]{whang2023data}
S.~E. Whang, Y.~Roh, H.~Song, and J.-G. Lee.
\newblock Data collection and quality challenges in deep learning: A data-centric ai perspective.
\newblock \emph{The VLDB Journal}, 32\penalty0 (4):\penalty0 791--813, 2023.

\bibitem[Wong et~al.(2023)Wong, Zheng, Valeri, Donghia, Anahtar, Omori, Li, Cubillos-Ruiz, Krishnan, Jin, et~al.]{wong2023discovery}
F.~Wong, E.~J. Zheng, J.~A. Valeri, N.~M. Donghia, M.~N. Anahtar, S.~Omori, A.~Li, A.~Cubillos-Ruiz, A.~Krishnan, W.~Jin, et~al.
\newblock Discovery of a structural class of antibiotics with explainable deep learning.
\newblock \emph{Nature}, pages 1--9, 2023.

\bibitem[Wright(2017)]{wright2017opportunities}
G.~D. Wright.
\newblock Opportunities for natural products in 21 st century antibiotic discovery.
\newblock \emph{Natural product reports}, 34\penalty0 (7):\penalty0 694--701, 2017.

\bibitem[Wyrsch et~al.(2022)Wyrsch, Nesporova, Tarabai, Jamborova, Bitar, Literak, Dolejska, and Djordjevic]{wyrsch2022urban}
E.~R. Wyrsch, K.~Nesporova, H.~Tarabai, I.~Jamborova, I.~Bitar, I.~Literak, M.~Dolejska, and S.~P. Djordjevic.
\newblock Urban wildlife crisis: Australian silver gull is a bystander host to widespread clinical antibiotic resistance.
\newblock \emph{Msystems}, 7\penalty0 (3):\penalty0 e00158--22, 2022.

\bibitem[Xiao et~al.(2009)Xiao, Tao, and Chen]{xiao2009optimal}
X.~Xiao, Y.~Tao, and M.~Chen.
\newblock Optimal random perturbation at multiple privacy levels.
\newblock \emph{Proceedings of the VLDB Endowment}, 2\penalty0 (1):\penalty0 814--825, 2009.

\bibitem[Xie et~al.(2018)Xie, Shen, and Zhao]{xie2018antibiotics}
W.-Y. Xie, Q.~Shen, and F.~Zhao.
\newblock Antibiotics and antibiotic resistance from animal manures to soil: a review.
\newblock \emph{European journal of soil science}, 69\penalty0 (1):\penalty0 181--195, 2018.

\bibitem[Yang et~al.(2020)Yang, Yu, You, Steinhardt, and Ma]{yang2020rethinking}
Z.~Yang, Y.~Yu, C.~You, J.~Steinhardt, and Y.~Ma.
\newblock Rethinking bias-variance trade-off for generalization of neural networks.
\newblock In \emph{International Conference on Machine Learning}, pages 10767--10777. PMLR, 2020.

\bibitem[Yelin et~al.(2019)Yelin, Snitser, Novich, Katz, Tal, Parizade, Chodick, Koren, Shalev, and Kishony]{yelin2019personal}
I.~Yelin, O.~Snitser, G.~Novich, R.~Katz, O.~Tal, M.~Parizade, G.~Chodick, G.~Koren, V.~Shalev, and R.~Kishony.
\newblock Personal clinical history predicts antibiotic resistance of urinary tract infections.
\newblock \emph{Nature medicine}, 25\penalty0 (7):\penalty0 1143--1152, 2019.

\bibitem[Youn et~al.(2022)Youn, Rai, and Tagkopoulos]{Youn2022}
J.~Youn, N.~Rai, and I.~Tagkopoulos.
\newblock Knowledge integration and decision support for accelerated discovery of antibiotic resistance genes.
\newblock \emph{Nature Communications}, 13:\penalty0 2360, 4 2022.
\newblock ISSN 2041-1723.
\newblock \doi{10.1038/s41467-022-29993-z}.

\bibitem[Zhang et~al.(2022)Zhang, Khosravi, Vahdati, Faghani, Nugen, Rassoulinejad-Mousavi, Moassefi, Jagtap, Singh, Rouzrokh, et~al.]{zhang2022mitigating}
K.~Zhang, B.~Khosravi, S.~Vahdati, S.~Faghani, F.~Nugen, S.~M. Rassoulinejad-Mousavi, M.~Moassefi, J.~M.~M. Jagtap, Y.~Singh, P.~Rouzrokh, et~al.
\newblock Mitigating bias in radiology machine learning: 2. model development.
\newblock \emph{Radiology: Artificial Intelligence}, 4\penalty0 (5):\penalty0 e220010, 2022.

\bibitem[Zhang et~al.(2015)Zhang, Ying, Pan, Liu, and Zhao]{zhang2015comprehensive}
Q.-Q. Zhang, G.-G. Ying, C.-G. Pan, Y.-S. Liu, and J.-L. Zhao.
\newblock Comprehensive evaluation of antibiotics emission and fate in the river basins of china: source analysis, multimedia modeling, and linkage to bacterial resistance.
\newblock \emph{Environmental science \& technology}, 49\penalty0 (11):\penalty0 6772--6782, 2015.

\bibitem[Zhao et~al.(2024)Zhao, Wu, Jiang, He, and Hu]{zhao2024causal}
W.~Zhao, J.~Wu, X.~Jiang, T.~He, and X.~Hu.
\newblock Causal-arg: a causality-guided framework for annotating properties of antibiotic resistance genes.
\newblock \emph{Bioinformatics}, 40\penalty0 (4):\penalty0 btae180, 2024.

\bibitem[Zhu et~al.(2013)Zhu, Johnson, Su, Qiao, Guo, Stedtfeld, Hashsham, and Tiedje]{zhu2013diverse}
Y.-G. Zhu, T.~A. Johnson, J.-Q. Su, M.~Qiao, G.-X. Guo, R.~D. Stedtfeld, S.~A. Hashsham, and J.~M. Tiedje.
\newblock Diverse and abundant antibiotic resistance genes in chinese swine farms.
\newblock \emph{Proceedings of the National Academy of Sciences}, 110\penalty0 (9):\penalty0 3435--3440, 2013.

\end{thebibliography}

\end{document}